\documentclass[review]{elsarticle}
\graphicspath{ {./figures/} }
\usepackage[hypertexnames=false]{hyperref}
\usepackage{float}
\usepackage{verbatim} 
\usepackage{apalike}
\usepackage{amsfonts}
\usepackage{algorithmicx}
\usepackage{amsmath}
\usepackage{amssymb}
\usepackage{algpseudocode}
\usepackage{algorithm}
\usepackage{booktabs}
\usepackage{tabularx}
\usepackage{placeins}
\usepackage{caption}
\floatstyle{plain} 
\restylefloat{figure}
\floatstyle{plaintop} 
\restylefloat{table}
\makeatletter
\def\ps@pprintTitle{%
  \let\@oddhead\@empty
  \let\@evenhead\@empty
  \let\@oddfoot\@empty
  \let\@evenfoot\@empty
}
\makeatother

\journal{Expert Systems with Applications}

\biboptions{authoryear}

\begin{document}
\begin{frontmatter}

\title{FedDP-PALD: Privacy-Preserving Federated Learning with a Latent Diffusion Model and Prototype Mixture Aggregation for Medical Data Synthesis}

\author[label1]{Md. Sajeebul Islam Sk.$^\dagger$}
\ead{mdsajeebulislamsk@mail.com}

\author[label2]{Khan Enaet Hossain$^\dagger$\corref{cor1}}
\ead{enaet09@yorku.ca}

\author[label3]{Md. Mehedi Hasan Shawon}
\ead{mshawon@umd.edu}

\fntext[ ]{$^\dagger$ These authors contributed equally to this work.}
\cortext[cor1]{Corresponding author.}
\address[label1]{Department of Computer Science and Engineering, BRAC University, Bangladesh}
\address[label2]{Department of Mathematics and Statistics, York University, Toronto, Canada}
\address[label3]{Department of Electrical and Computer Engineering, University of Maryland, College Park, Maryland, USA}

\begin{abstract}
Medical images and physiological signals provide valuable information for accurate diagnosis. Developing diagnostic models often requires patient data from multiple institutions, although strict privacy regulations limit the sharing of sensitive clinical records. Federated learning enables multiple hospitals to train a shared model without exchanging raw data. However, existing methods face two problems: the information exchanged during training can reveal whether a patient's data were used, and synthetic data meant to replace real records often fail to preserve their predictive structure, which limits clinical use. To address this issue, we propose FedDP-PALD, a privacy-preserving federated latent diffusion framework for multimodal medical data synthesis under formal privacy guarantees. It jointly processes chest X-ray images and electrocardiogram (ECG) signals through gated multi-head attention with modality-availability masks, remaining effective even when a modality is missing. We also introduce Differentially Private Prototype Mixture Aggregation (DP-PMA), which clips class-level latent prototypes and adds calibrated Gaussian noise before combining them on the server to maintain $(\epsilon, \delta)$ differential privacy. We evaluate FedDP-PALD on PneumoniaMNIST, ChestMNIST, and MIT-BIH datasets, where differential privacy reduced summary-level attack AUROC from 0.6229 $\pm$ 0.0026 to between 0.5016 and 0.5093 for privacy budgets from $\epsilon = 1$ to $\epsilon = 8$. On the test data, synthetic-latent training achieved an F1 score of 0.8993 $\pm$ 0.0006 and an AUROC of 0.9057 $\pm$ 0.0503, close to the 0.9747 $\pm$ 0.0132 real-latent training. These results show that FedDP-PALD generates private synthetic representations that preserve useful decision performance while strongly resisting membership inference.
\end{abstract}

\begin{keyword}
Federated learning \sep Differential privacy \sep Diffusion models \sep Prototype mixture aggregation \sep Multimodal medical data synthesis
\end{keyword}

\end{frontmatter}

\section{Introduction}
\label{sec:introduction}

\noindent Medical decisions increasingly rely on information from different sources. Images reveal anatomical structure, whereas physiological signals capture how the body functions over time. Using both modalities can provide a richer view of disease than either one alone. Medical data are commonly distributed across hospitals, laboratories, and monitoring centres, and their centralized collection is often restricted by privacy requirements, institutional policies, and data-governance constraints. These restrictions limit the amount and diversity of data available to conventional machine-learning systems and make multi-institutional model development difficult \cite{rieke2020future, sheller2020federated}. Federated learning enables a practical alternative. Each institution trains a local model and shares model-related information with a coordinating server rather than transferring raw clinical records \cite{mcmahan2017communication}. This arrangement has made collaborative medical learning more feasible, particularly when data are distributed across sites with different class distributions, sample sizes, and acquisition procedures \cite{sheller2020federated, khowaja2025selffed}. Still, federated learning is not private by default. Gradients, parameters, predictions, and latent representations may retain traces of individual training samples. Membership inference attacks exploit these traces to estimate whether a particular record was used during training \cite{shokri2017membership, ye2022enhanced}. In a medical modality, even that binary disclosure can be sensitive.

\noindent Differential privacy provides a mathematical framework for limiting the influence of an individual sample on a released output. It bounds the change in the output distribution when one training record is added or removed \cite{dwork2006calibrating, abadi2016deep}. Differentially private federated learning typically injects calibrated noise into gradients, model parameters, or server-side aggregates. These approaches reduce privacy leakage, but repeated perturbation during optimization can slow convergence and lower predictive performance, especially under non-identically distributed clients. Clinical sites differ in disease prevalence, scanner type, signal quality, and available modalities. Prototype-based federated learning is one alternative to gradient-space exchange. Instead of sharing full representations or gradients, clients communicate compact class-level summaries. FedProto uses class prototypes to support learning under client heterogeneity \cite{tan2022fedproto}, and related studies treat private prototypes as lower-dimensional summaries of sensitive embeddings \cite{ribero2022federating, wahdany2025differentially}. A prototype is still data-dependent. Released without protection, it can reveal information about the samples it was computed from. A single class centroid is also coarse when a class contains several latent subgroups, and prototype order rarely matches across clients, so naive averaging may combine unrelated components.

\noindent Synthetic data generation is often proposed to reduce dependence on real records. Diffusion models are attractive because they represent complex distributions through iterative noising and denoising. Latent diffusion makes this tractable by operating in a compact feature space rather than on high-dimensional images or signals. Private latent diffusion has therefore drawn attention in medical and general data synthesis \cite{lyu2024dpldm, daum2024private}. Synthetic data are not automatically safe. A generator can reproduce local neighbourhood structure, leak membership information, or produce samples that look plausible yet fail to support reliable prediction. Privacy, utility, and fidelity should therefore be examined together \cite{adams2025fidelity, hernandez2025evaluation}. Multimodal federated learning adds a further challenge. Clients may not hold the same modalities, and some observations are incomplete. Prior work handles this through reconstruction, contrastive alignment, prototype matching, and missing-modality completion \cite{bao2024prototype, le2025crossmodal, poudel2025feature}. FlexiD-Fuse uses diffusion to fuse a varying number of medical images \cite{xu2025flexid}, and SelfFed targets heterogeneous medical federated learning and label scarcity through decentralized pretraining and adaptive aggregation \cite{khowaja2025selffed}. FlexiD-Fuse performs centralized image fusion, and SelfFed studies federated classification rather than private synthetic latent generation. Federated medical learning, missing-modality handling, private prototype release, and latent diffusion have mostly been studied in isolation. Existing prototype methods may not preserve multiple class-conditional components across heterogeneous clients, and private generative studies often report utility without examining ranking loss, minority-class coverage, or repeated membership attacks. There is little evidence on whether a diffusion model trained from a compact, differentially private prototype mixture can generate useful synthetic medical representations without storing client-level latent vectors. We study these gaps and focus on four research questions.

\begin{itemize}
\item RQ1. Can modality-aware gated fusion learn useful class-aligned chest X-ray and ECG representations under heterogeneous federated training while remaining effective when one modality is unavailable?
\item RQ2. Can DP-PMA reduce membership leakage from released latent summaries across different privacy budgets without removing the class structure needed for downstream learning?
\item RQ3. Can a latent diffusion model trained only from the private prototype mixture generate synthetic representations with utility close to that of real latent representations?
\item RQ4. How do privacy, predictive utility, ranking performance, and distributional fidelity change across privacy budgets, and which failure modes remain after synthesis?
\end{itemize}

\noindent To address this gap and meet these goals, we propose FedDP-PALD, a privacy-preserving federated latent diffusion framework with Prototype Mixture Aggregation for medical data synthesis. FedDP-PALD encodes chest X-ray images and electrocardiogram (ECG) signals with separate encoders and fuses them through gated multi-head attention with modality-availability masks and modality dropout, so the model still operates when one modality is missing. After federated training, each client releases differentially private class-conditional prototype mixtures rather than individual latent vectors. We introduce Differentially Private Prototype Mixture Aggregation (DP-PMA) to align and merge these mixtures on the server under a formal $(\epsilon,\delta)$ guarantee. A class-conditional latent diffusion model is then trained only on the released mixture, without storing client latents, and its samples form synthetic training sets for downstream diagnosis. The main contributions of this study are as follows.

\begin{itemize}
\item We develop FedDP-PALD, a federated framework that combines modality-specific encoders, gated multi-head attention, availability masks, and modality dropout for class-aligned chest X-ray and ECG representation learning.
\item We introduce DP-PMA, which releases a class-conditional prototype mixture under $(\epsilon,\delta)$ differential privacy instead of raw latent vectors, preserving multiple intra-class components across heterogeneous clients.
\item We design a latent synthesis stage in which a class-conditional diffusion model is trained from the private prototype mixture without storing or replaying individual client latents.
\item We evaluate the framework under a joint privacy-utility-fidelity protocol spanning membership inference, disclosure and duplicate tests, train-synthetic-test-real classification, privacy-budget sweeps, missing-modality tests, and ablations.
\end{itemize}

\section{Related Work}
\label{sec:related}

Neural learning has been applied across medical imaging, ECG analysis, speech-based inference, and computational modelling of complex fluid systems \cite{islam2026vlm, islam2025foundationalecg, islam2024bangla, ISLAM2026129418, islam2025artificial, rana2025neural}.

\subsection{Federated Learning for Medical Data}

\noindent Federated learning enables multiple institutions to train a shared model while retaining raw records at their local sites \cite{mcmahan2017communication}. This structure has clear relevance to medicine, where patient confidentiality, institutional governance, and data-management policies restrict centralized data collection. Studies in medical imaging have shown that collaboration across institutions can be achieved without transferring complete patient datasets \cite{rieke2020future,sheller2020federated}. Large-scale reviews identify non-IID data, communication cost, client participation, and information leakage as continuing research challenges \cite{kairouz2021advances}. Differences among participating sites remain a central concern. Hospitals can vary in sample size, disease prevalence, scanner configuration, patient population, and annotation practice. Under these conditions, FedAvg may place greater influence on large clients and provide weaker performance for less representative sites. FedProx constrains local updates through proximal regularization \cite{li2020fedprox}. SelfFed instead combines decentralized pretraining with federated fine-tuning to improve learning from heterogeneous medical images with limited labels \cite{khowaja2025selffed}. Their principal focus is predictive learning under statistical heterogeneity. Private latent release and downstream generative evaluation fall outside their main scope. Data locality also provides no formal bound on information leakage. Parameters, gradients, confidence scores, and intermediate representations remain dependent on local training records. Membership inference attacks exploit behavioural differences between samples seen during training and held-out samples \cite{shokri2017membership,ye2022enhanced}. Within a medical study, identifying membership in a disease-specific cohort may itself disclose sensitive information. A federated architecture therefore reduces direct data movement, while an additional privacy mechanism is required when released information needs a measurable guarantee.

\subsection{Multimodal Federated Learning with Incomplete Inputs}

\noindent Clinical AI frequently draws information from several data forms. MRI-based vision-language models combine imaging features with textual evidence for diagnostic interpretation \cite{islam2026vlm}, while FoundationalECGNet combines wavelet processing, graph attention, convolutional attention, and temporal transformers for cardiac analysis \cite{islam2025foundationalecg}. These studies illustrate how modality-specific encoders can capture distinct anatomical and physiological patterns. Multimodal federated learning extends this principle across decentralized clients. FedMEKT transfers joint embedding knowledge through semi-supervised multimodal autoencoders and knowledge distillation \cite{fedmekt2025}. Bao et al.\ introduce prototype masks and prototype-guided objectives for clients with missing modality information \cite{bao2024prototype}. MFCPL uses complete prototypes, cross-modal regularization, and contrastive learning under severely incomplete modality modalitys \cite{le2025crossmodal}. FedMEPD adopts modality-specific encoders, a partially personalized fusion decoder, and server-derived anchors for multimodal brain-tumour segmentation \cite{fedmepd2025}. Feature-imputation networks provide another strategy by estimating absent representations before fusion \cite{poudel2025feature}. FedDP-PALD introduces greater emphasis on private latent release after federated fusion. Its fusion module uses modality-specific encoders, gated multi-head attention, availability masks, and modality dropout. The diffusion stage also serves a different purpose. FlexiD-Fuse applies diffusion to combine several medical images into one fused image \cite{xu2025flexid}; FedDP-PALD applies latent diffusion to create synthetic feature representations from a released private prototype distribution.

\subsection{Prototype-Based Federated Learning and Differential Privacy}

\noindent Prototype-based federated learning replaces the exchange of high-dimensional models with compact class-level representations. FedProto computes local prototypes, combines them at the server, and uses the resulting global prototypes to guide client learning under data and model heterogeneity \cite{tan2022fedproto}. This representation-level exchange can lower communication demand and support knowledge transfer among clients with different local architectures. Compactness alone gives no formal protection. A class prototype is derived from private embeddings, and its release may expose information about nearby observations when local class counts are small. Differential privacy addresses this risk by bounding the influence of one record on the probability distribution of a released output \cite{dwork2006calibrating}. DP-SGD clips per-sample gradients and injects Gaussian noise during iterative optimization \cite{abadi2016deep}. Output perturbation follows a different route: noise is added directly to a released statistic according to its sensitivity. The analytic Gaussian mechanism provides a precise calibration for this modality \cite{balle2018improving}. Ribero et al.\ use differentially private prototypes to support federated recommendation without collecting individual preference records \cite{ribero2022federating}. Wahdany et al.\ study private prototypes under imbalanced transfer learning and examine the relation between privacy noise and minority-class performance \cite{wahdany2025differentially}. These works establish the feasibility of private prototype exchange. Yet a single centroid may compress meaningful within-class variation. Several prototypes can preserve richer geometry, though their component indices have no shared semantic order across clients. Combining components solely by index may therefore join distant latent regions. DP-PMA represents each class through several fixed-shard prototypes. Client representations are norm-clipped before prototype construction, and Gaussian perturbation is calibrated under replace-one adjacency. A minimum-cost assignment subsequently matches related mixture components across clients before server aggregation. The formal $(\epsilon,\delta)$ privacy protection applies to the released prototype means. It does not extend to the preceding federated model updates.

\subsection{Private Synthetic Data and Latent Diffusion}

\noindent Synthetic medical data are studied as a means of supporting analysis when access to real records is restricted. Torfi et al.\ combine convolutional autoencoders, generative adversarial networks, and R'enyi differential privacy to generate medical data containing continuous and discrete variables \cite{torfi2022dpgan}. Their evaluation considers statistical similarity, downstream learning, and privacy-budget effects. This work illustrates an early effort to assess medical generation through both privacy and task utility. Diffusion models provide a later generative approach based on gradual noise injection followed by learned denoising \cite{ho2020denoising}. Latent diffusion transfers this process to a compressed representation space, reducing the computation required for high-dimensional data \cite{rombach2022high}. DP-LDM applies differential privacy while adapting latent-diffusion components and studies the associated privacy-quality relation \cite{lyu2024dpldm}. Daum et al.\ apply private latent diffusion to three-dimensional cardiac-image synthesis and report lower image quality and controllability under stronger protection \cite{daum2024private}. Medical diffusion studies also examine downstream value. Niehues et al.\ compare latent diffusion with GAN-based histopathology generation, evaluate memorization, and measure classification performance after synthetic-data augmentation \cite{niehues2024histology}. Their findings show that generative quality is assessed beyond visual appearance. A sample can appear plausible while representing only a narrow region of the real distribution. Equally, low nearest-neighbour similarity provides limited evidence about downstream discrimination. Recent evaluation studies consequently separate privacy, utility, and fidelity \cite{adams2025fidelity,hernandez2025evaluation}. Train-synthetic-test-real experiments estimate whether possible data retain decision-related information. Membership attacks, neighbour distances, and duplicate tests examine disclosure risk. Distributional distances and class-conditional coverage reveal whether the generator represents the breadth of the source space. Existing studies introduce protection during generator training, train latent models from real embeddings, and assess generation after direct access to private training data. We use a different sequence FedDP-PALD. The server receives an assignment-aligned private prototype mixture, and the diffusion model learns from samples drawn from that released distribution. Individual client latents are absent from diffusion training. The literature therefore provides strong foundations for each stage of the proposed framework, yet limited work connects modality-aware federated representation learning, private mixture release, component alignment, and downstream latent generation within one evaluated pipeline. We combine these stages and examine the resulting system through membership inference, nearest-neighbour disclosure, duplicate analysis, missing-modality tests, distributional fidelity, and train-synthetic-test-real classification. The contribution lies in the interaction among these stages and in the joint analysis of privacy protection, decision utility, ranking performance, and latent coverage.

\section{Methodology}
\label{sec:method}

\begin{figure}[!htbp]
    \centering
    \makebox[\textwidth][c]{%
        \includegraphics[
            width=\textwidth,
            keepaspectratio
        ]{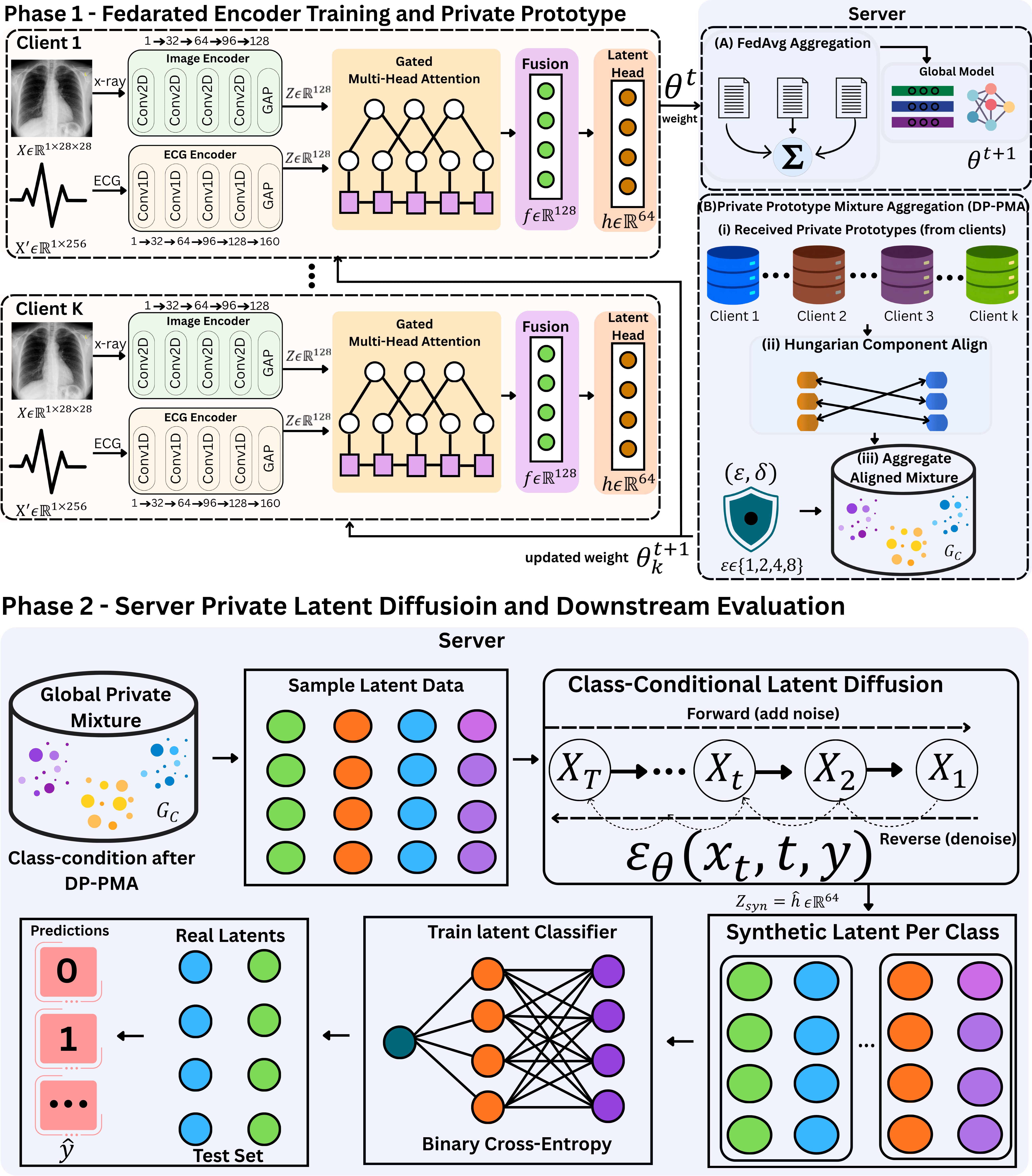}
    }
    \caption{Detail workflow of the proposed FedDP-PALD framework. Phase~1 performs federated multimodal representation learning, FedAvg-based model updating, differentially private prototype release, Hungarian component alignment, and global private mixture aggregation. Phase~2 trains a class-conditional latent diffusion model from samples drawn from the global private mixture, generates synthetic 64-dimensional latent representations, and evaluates a downstream classifier on held-out real latent data.}
    \label{fig:feddp_workflow}
\end{figure}

\noindent We develop FedDP-PALD as a two-phase framework. In Phase~1, the model learns class-aligned chest X-ray and ECG representations through
federated multimodal encoder training and releases differentially private
prototype mixtures. In Phase~2, the server constructs a global private
mixture, trains a class-conditional latent diffusion model, and generates
synthetic latent representations for downstream classification. The
detail workflow is presented in
Figure~\ref{fig:feddp_workflow}.

\subsection{Problem Formulation}
\label{subsec:problem_formulation}

Let $\mathcal{K}=\{1,\ldots,K\}$ denote a federation of $K$ clients. Client $k$ holds a local dataset

\begin{equation}
\mathcal{D}_{k}
=
\left\{
\left(
\mathbf{x}^{I}_{k,i},
\mathbf{x}^{E}_{k,i},
y_{k,i},
\mathbf{m}_{k,i}
\right)
\right\}_{i=1}^{n_k},
\label{eq:local_dataset}
\end{equation}

\noindent where $\mathbf{x}^{I}_{k,i}\in\mathbb{R}^{1\times28\times28}$ is a chest X-ray image, $\mathbf{x}^{E}_{k,i}\in\mathbb{R}^{1\times256}$ is
an ECG signal, $y_{k,i}\in\{0,1\}$ is the harmonized binary label, and $\mathbf{m}_{k,i}=[m^{I}_{k,i},m^{E}_{k,i}]\in\{0,1\}^{2}$ records the
availability of the two modalities. Federated representation learning seeks the global parameters without transferring raw X-ray images or ECG signals.

\begin{equation}
\boldsymbol{\theta}^{t}
=
\arg\min_{\boldsymbol{\theta}}
\sum_{k=1}^{K}
\frac{n_k}{\sum_{j=1}^{K}n_j}
\mathcal{L}_{k}(\boldsymbol{\theta}),
\label{eq:global_objective}
\end{equation}

\noindent where $\boldsymbol{\theta}$ denotes the complete trainable parameter set of FedDP-PALD, including the image encoder, ECG encoder, gated multi-head attention module, fusion layer, latent head, and classifier.
$\boldsymbol{\theta}^{t}$ represents the optimized global model
parameters. The symbol $K$ is the number of participating clients,
$n_k$ is the number of training samples held by client $k$, and
$\sum_{j=1}^{K}n_j$ is the total number of training samples across all
clients. The factor $\frac{n_k}{\sum_{j=1}^{K}n_j}$ assigns a sample-size-dependent contribution to client $k$ and $\mathcal{L}_{k}(\boldsymbol{\theta})$ denotes the local classification loss evaluated on client $k$. After federated training, the differentially private client prototypes are aligned and aggregated to construct the global class-conditional private mixture

\begin{equation}
{{G}}_{c}
=
\sum_{r=1}^{R_p}
\widetilde{\pi}_{c,r}
\mathcal{N}
\left(
\widetilde{\boldsymbol{\mu}}_{c,r},
\operatorname{diag}
\left(
\widetilde{\boldsymbol{\Sigma}}_{c,r}
\right)
\right),
\qquad c\in\{0,1\},
\label{eq:private_distribution}
\end{equation}

\noindent where $R_p$ is the number of aligned mixture components,
$\widetilde{\pi}_{c,r}$ is the weight of component $r$ for class $c$,
$\widetilde{\boldsymbol{\mu}}_{c,r}$ is its differentially private mean,
and $\widetilde{\boldsymbol{\Sigma}}_{c,r}$ is its diagonal covariance, respectively. Then the server samples latent vectors from this mixture and uses these samples to train the class-conditional latent diffusion
model. After diffusion training, the reverse denoising process produces
the final synthetic latent representations:

\begin{equation}
\mathbf{x}_{0}\sim{{G}}_{c},
\qquad c\in\{0,1\},
\label{eq:mixture_sampling}
\end{equation}

\begin{equation}
\mathbf{z}_{\mathrm{syn}}
=
\widehat{\mathbf{x}}_{0}
\in\mathbb{R}^{64}.
\label{eq:synthetic_representation}
\end{equation}

\subsection{Datasets and Data Preprocessing}
\label{subsec:data_preprocessing}

\noindent We use three publicly available medical datasets in this study.
PneumoniaMNIST provides the primary chest X-ray observations,
ChestMNIST is used for auxiliary image-encoder pretraining, and the MIT-BIH Arrhythmia Database provides ECG signals. PneumoniaMNIST and ChestMNIST contain standardized $28\times28$ grayscale images \cite{yang2023medmnist}, whereas MIT-BIH contains annotated ECG recordings \cite{goldberger2000physionet}. Table~\ref{tab:datasets} represents datasets modalities and access sources.

\begin{table}[t]
\centering
\caption{Datasets used in the proposed framework.}
\label{tab:datasets}
\small
\begin{tabularx}{\columnwidth}{l l X l}
\toprule
\textbf{Dataset} & \textbf{Modality} & \textbf{Role} & \textbf{Access} \\
\midrule
PneumoniaMNIST
& Chest X-ray
& image input
& \cite{yang2023medmnist} \\

ChestMNIST
& Chest X-ray
& 14-label encoder
& \cite{yang2023medmnist} \\

MIT-BIH
& ECG
& signal input
& \cite{goldberger2000physionet} \\
\bottomrule
\end{tabularx}
\end{table}

\noindent Each chest X-ray was converted to floating-point format,
rescaled to $[0,1]$, and then standardized using only the mean and
standard deviation estimated from the training images. The resulting
normalized image input is defined in Eq.~\eqref{eq:image_preprocessing}:

\begin{equation}
\widehat{\mathbf{x}}^{I}
=
\frac{\mathbf{x}^{I}/255-\mu_I}{\sigma_I},
\label{eq:image_preprocessing}
\end{equation}

\noindent where $\mathbf{x}^{I}$ denotes the original image, while $\mu_I$ and
$\sigma_I$ are the mean and standard deviation computed from the
PneumoniaMNIST training period. For MIT-BIH, we performed the split at the record level, using 70\% of the records for training, 15\% for validation, and 15\% for testing. This separation prevents signals from the same ECG recording from appearing in more than one period. We selected the MLII channel
whenever it was available, followed by MLI, and otherwise used the first
recorded channel. Around each annotated signal, we extracted a 256-sample
window containing 96 samples before and 160 samples after the annotated
location. Normal signals were assigned to class 0, and abnormal signals were assigned to class 1. Each ECG segment was first centered and scaled using its own sample statistics. Then we standardized the resulting signal with the mean and standard deviation estimated from the ECG training period, as shown in Eq.~\eqref{eq:ecg_preprocessing}:

\begin{equation}
\widetilde{\mathbf{x}}^{E}
=
\frac{
\mathbf{x}^{E}
-
\operatorname{mean}(\mathbf{x}^{E})
}{
\max\!\left\{
\operatorname{std}(\mathbf{x}^{E}),
10^{-6}
\right\}
},
\qquad
\widehat{\mathbf{x}}^{E}
=
\frac{
\widetilde{\mathbf{x}}^{E}
-
\mu_E
}{
\sigma_E
}.
\label{eq:ecg_preprocessing}
\end{equation}

\noindent Here, $\mathbf{x}^{E}$ is the extracted ECG signal, $\widetilde{\mathbf{x}}^{E}$ is the segment normalization, $\mu_E$ and $\sigma_E$ are calculated from the ECG training period.

\subsection{Federated Data Construction}
\label{subsec:federated_data}

\noindent The preprocessed data are distributed across $K=5$ clients. PneumoniaMNIST images are partitioned class-wise using a Dirichlet distribution \cite{Tomarchio2025} with a concentration parameter $\alpha=0.5$ and MIT-BIH signals are assigned at the record level so that signals from the same ECG
record remain within a single client. Since the X-ray and ECG data from independent datasets, we combined them as class-aligned pseudo-pairs. For each class $c\in\{0,1\}$, an X-ray image is paired with an ECG signal carrying the same binary label, as defined in Eq.~\eqref{eq:class_aligned_pairs}:

\begin{equation}
\mathcal{P}_{k}^{c}
=
\left\{
\left(
\widehat{\mathbf{x}}^{I}_{k,i},
\widehat{\mathbf{x}}^{E}_{k,\pi_c(i)},
c
\right)
:
y^{I}_{k,i}
=
y^{E}_{k,\pi_c(i)}
=
c
\right\},
\label{eq:class_aligned_pairs}
\end{equation}

\noindent where $\pi_c(i)$ denotes the class-specific pairing index, each modality is
independently dropped during local training with probability $p_m=0.25$. At least one modality is retained for every observation. The associated mask is defined in Eq.~\eqref{eq:modality_mask}:

\begin{equation}
\mathbf{m}_{k,i}
=
\left[
m^{I}_{k,i},
m^{E}_{k,i}
\right],
\qquad
m^{I}_{k,i}+m^{E}_{k,i}\geq1,
\label{eq:modality_mask}
\end{equation}

\noindent where $m^{I}_{k,i}$ and $m^{E}_{k,i}$ denote the availability of the X-ray and ECG modalities, respectively.

\subsection{Modality-Specific Feature Extraction}
\label{subsec:feature_extraction}
\noindent We use a weighted binary cross-entropy to balance the ChestMNIST labels \cite{yang2023medmnist}:

\begin{equation}
\mathcal{L}_{\mathrm{pre}}
=
-\frac{1}{NL}
\sum_{i=1}^{N}
\sum_{\ell=1}^{L}
\left[
w_{\ell}y_{i,\ell}\log \sigma(s_{i,\ell})
+
(1-y_{i,\ell})
\log\!\left(1-\sigma(s_{i,\ell})\right)
\right]
\label{eq:pretraining_loss}
\end{equation}

\noindent In Eq.~\eqref{eq:pretraining_loss}, $N$ is the number of training samples, $L=14$ is the number of labels, $s_{i,\ell}$ is the logit for label $\ell$, and $w_{\ell}$ is the positive-class weight calculated from the corresponding negative and positive training counts, and $w_{\ell}$ is the class-ratio. After pretraining, the classification head is removed, and only the encoder parameters are transferred to the federated model. We used a two-dimensional convolutional architecture \cite{lecun1998gradient} in the chest X-ray encoder with channel progression. Each block contains a $3\times3$ convolution, batch normalization \cite{ioffe2015batch}, and GELU activation \cite{hendrycks2016gelu}. Global average pooling \cite{lin2014network}, linear projection, and layer normalization \cite{ba2016layer} produce the two 128-dimensional modality representations:

\begin{equation}
\mathbf{z}^{I}
=
E_I(\widehat{\mathbf{x}}^{I})
\in\mathbb{R}^{128},
\qquad
\mathbf{z}^{E}
=
E_E(\widehat{\mathbf{x}}^{E})
\in\mathbb{R}^{128}.
\label{eq:modality_embeddings}
\end{equation}

\noindent Here, $E_I(\cdot)$ and $E_E(\cdot)$ denote the image and ECG encoders, $\widehat{\mathbf{x}}^{I}$ and
$\widehat{\mathbf{x}}^{E}$ are the preprocessed inputs defined in Eqs.~\eqref{eq:image_preprocessing} and~\eqref{eq:ecg_preprocessing}. The separate pathways retain modality-specific spatial and temporal
patterns before cross-modal fusion.

\subsection{Federated Multimodal Representation Learning}
\label{subsec:federated_representation}

\noindent The image and ECG embeddings in Eq.~\eqref{eq:modality_embeddings} are projected into a common 128-dimensional space and arranged as two modality tokens:

\begin{equation}
\mathbf{T}_{i}
=
\begin{bmatrix}
\mathbf{W}_{I}\mathbf{z}^{I}_{i}+\mathbf{b}_{I}\\
\mathbf{W}_{E}\mathbf{z}^{E}_{i}+\mathbf{b}_{E}
\end{bmatrix}
\in\mathbb{R}^{2\times128}.
\label{eq:modality_tokens}
\end{equation}

\noindent In Eq.~\eqref{eq:modality_tokens}, $\mathbf{W}_{I}$ and $\mathbf{W}_{E}$ are learned modality projections, and $\mathbf{b}_{I}$ and $\mathbf{b}_{E}$ are their bias vectors. Four-head self-attention, based on the scaled dot-product attention mechanism of Vaswani et al.\ \cite{vaswani2017attention}, models interaction between the two tokens. The availability mask suppresses a missing modality, and the proposed learned gate controls the attention output and the original token:

\begin{equation}
\mathbf{H}_{i}
=
\operatorname{MHA}
(\mathbf{T}_{i},\mathbf{m}_{i}),
\qquad
\mathbf{g}_{i}
=
\sigma
\left[
\mathbf{W}_{g,2}
\operatorname{GELU}
\left(
\mathbf{W}_{g,1}\mathbf{T}_{i}
+
\mathbf{b}_{g,1}
\right)
+
\mathbf{b}_{g,2}
\right]
\end{equation}

\begin{equation}
\mathbf{U}_{i}
=
\mathbf{g}_{i}\odot\mathbf{H}_{i}
+
(1-\mathbf{g}_{i})\odot\mathbf{T}_{i}.
\label{eq:gated_attention}
\end{equation}

\noindent Here, $\mathbf{H}_{i}$ is the multi-head attention output, $\mathbf{g}_{i}\in(0,1)^{2\times1}$ contains the learned token gates and $\odot$ denotes element-wise multiplication. Following linear transformation, layer normalization \cite{ba2016layer}, GELU activation \cite{hendrycks2016gelu}, and dropout, the available tokens are pooled as

\begin{equation}
\mathbf{f}_{i}
=
\frac{
\sum_{q=1}^{2}
m_{i,q}\mathbf{u}_{i,q}
}{
\max
\left(
\sum_{q=1}^{2}m_{i,q},
1
\right)
}
\in\mathbb{R}^{128}.
\label{eq:masked_fusion}
\end{equation}

\noindent In Eq.~\eqref{eq:masked_fusion}, $\mathbf{u}_{i,q}$ is the transformed token of modality $q$, and $m_{i,q}$ represents whether the modality is available. The denominator normalizes the fused representation by the number of available modalities. The latent head projects $\mathbf{f}_{i}$ into a shared 64-dimensional representation, after which a linear classifier produces the local class probabilities:

\begin{equation}
\mathbf{h}_{i}
=
H_{\phi}(\mathbf{f}_{i})
\in\mathbb{R}^{64},
\qquad
\widehat{\mathbf{p}}_{i}
=
\operatorname{softmax}
\left(
\mathbf{W}_{c}\mathbf{h}_{i}
+
\mathbf{b}_{c}
\right),
\qquad
\widehat{y}_{i}
=
\arg\max_{c\in\{0,1\}}
\widehat{p}_{i,c}.
\label{eq:latent_prediction}
\end{equation}

\noindent Here, $H_{\phi}$ denotes the latent projection head, $\widehat{\mathbf{p}}_{i}$ contains the two class probabilities, and $\widehat{y}_{i}$ is the predicted label. At round $t$, the server provides the current global parameters $\boldsymbol{\theta}^{(t)}$ to all clients. Each client performs $E$ local epochs using cross-entropy loss and returns its updated parameters. Following the FedAvg method \cite{mcmahan2017communication}, the local objective and global update are written as:

\begin{equation}
\mathcal{L}_{k}^{\mathrm{cls}}
=
-\frac{1}{n_k}
\sum_{i=1}^{n_k}
\log
\widehat{p}_{k,i,y_{k,i}},
\qquad
\boldsymbol{\theta}^{(t+1)}
=
\sum_{k=1}^{K}
\frac{n_k}{\sum_{j=1}^{K}n_j}
\boldsymbol{\theta}_{k}^{(t+1)}.
\label{eq:local_loss_fedavg}
\end{equation}

\noindent In Eq.~\eqref{eq:local_loss_fedavg}, $n_k$ is the number of training samples at client $k$, $\mathcal{L}_{k}^{\mathrm{cls}}$ is its local classification loss, and $\boldsymbol{\theta}_{k}^{(t+1)}$ denotes its locally updated model. The sample-size coefficient assigns each client a contribution proportional to its local dataset. The implementation uses four local epochs and eight communication rounds. The exchanged parameters include the image encoder, ECG encoder, gated attention module, fusion layer, latent head, and classifier. Differential privacy is not applied to these model parameters.

\subsection{Differentially Private Prototype Mixture Aggregation (DP-PMA)}
\label{subsec:dppma}

\noindent After federated training, the global encoder produces $\mathbf{h}_{k,i}\in\mathbb{R}^{64}$ for each client observation. DP-PMA is then applied locally. Each class-specific latent vector is clipped to the radius $C$ and assigned to one of $R_p=4$ fixed shards:

\begin{equation}
\overline{\mathbf{h}}_{k,i}
=
\mathbf{h}_{k,i}
\min
\left(
1,
\frac{C}{
\max
\left(
\|\mathbf{h}_{k,i}\|_2,
10^{-12}
\right)
}
\right),
\qquad
\boldsymbol{\mu}_{k,c,r}
=
\frac{1}{n_{k,c,r}}
\sum_{i\in\mathcal{S}_{k,c,r}}
\overline{\mathbf{h}}_{k,i}.
\label{eq:clip_and_mean}
\end{equation}

\noindent Here, $\overline{\mathbf{h}}_{k,i}$ is the clipped latent vector, $\mathcal{S}_{k,c,r}$ is shard $r$ of class $c$, and $n_{k,c,r}$ is the number of observations in that shard. In equation~\eqref{eq:clip_and_mean}, we define the proposed fixed-shard prototype construction. Under fixed-size replace-one adjacency, the triangle inequality gives the shard-mean sensitivity.

\begin{equation}
\Delta_{2,k,c,r}
=
\frac{2C}{n_{k,c,r}}.
\label{eq:prototype_sensitivity}
\end{equation}

\noindent The Gaussian mechanism was introduced as a differential-privacy mechanism by Dwork et al. \cite{dwork2006calibrating}. We use the analytic calibration of Balle and Wang \cite{balle2018improving} to release the private shard mean:

\begin{equation}
\widetilde{\boldsymbol{\mu}}_{k,c,r}
=
\boldsymbol{\mu}_{k,c,r}
+
\boldsymbol{\xi}_{k,c,r},
\qquad
\boldsymbol{\xi}_{k,c,r}
\sim
\mathcal{N}
\left(
\mathbf{0},
\sigma_{k,c,r}^{2}\mathbf{I}
\right).
\label{eq:private_prototype}
\end{equation}

\noindent The standard deviation $\sigma_{k,c,r}$ is calibrated from the sensitivity in Eq.~\eqref{eq:prototype_sensitivity} by solving the analytic Gaussian condition \cite{balle2018improving}:

\begin{equation}
\delta
=
\Phi
\left(
\frac{\Delta_2}{2\sigma}
-
\frac{\varepsilon\sigma}{\Delta_2}
\right)
-
e^{\varepsilon}
\Phi
\left(
-\frac{\Delta_2}{2\sigma}
-
\frac{\varepsilon\sigma}{\Delta_2}
\right),
\label{eq:analytic_gaussian}
\end{equation}

\noindent where $\Phi(\cdot)$ is the cumulative distribution function of the standard Gaussian distribution and $(\varepsilon,\delta)$ is the selected privacy budget. The component index has no shared semantic order across clients, so we use the Hungarian assignment method to align them \cite{kuhn1955hungarian}:

\begin{equation}
\rho_{k,c}^{\star}
=
\arg\min_{\rho\in\mathfrak{S}_{R_p}}
\sum_{r=1}^{R_p}
\left\|
\widetilde{\boldsymbol{\mu}}_{k,c,r}
-
\widetilde{\boldsymbol{\mu}}_
{k_c^{\star},c,\rho(r)}
\right\|_2.
\label{eq:hungarian_alignment}
\end{equation}

\noindent In Eq.~\eqref{eq:hungarian_alignment}, $k_c^{\star}$ is the reference client with the largest class-$c$ contribution, $\mathfrak{S}_{R_p}$ is the set of component permutations and $\rho_{k,c}^{\star}$ is the minimum-cost assignment. For aligned component $r$:

\begin{equation}
a_{k,c,r}
=
n_{k,c}
\pi_{k,c,\rho_{k,c}^{-1}(r)},
\qquad
A_{c,r}
=
\sum_{k=1}^{K}a_{k,c,r}.
\label{eq:component_mass}
\end{equation}

\noindent The global private parameters are then obtained through the proposed sample-mass-weighted aggregation:

\begin{align}
\widetilde{\boldsymbol{\mu}}_{c,r}
&=
\frac{1}{A_{c,r}}
\sum_{k=1}^{K}
a_{k,c,r}
\widetilde{\boldsymbol{\mu}}_
{k,c,\rho_{k,c}^{-1}(r)},
\nonumber\\
\widetilde{\boldsymbol{\Sigma}}_{c,r}
&=
\max
\left[
\frac{1}{A_{c,r}}
\sum_{k=1}^{K}
a_{k,c,r}
\left(
\boldsymbol{\Sigma}_{c}
+
\widetilde{\boldsymbol{\mu}}_
{k,c,\rho_{k,c}^{-1}(r)}^{\odot2}
\right)
-
\widetilde{\boldsymbol{\mu}}_{c,r}^{\odot2},
10^{-4}
\right],
\nonumber\\
\widetilde{\pi}_{c,r}
&=
\frac{A_{c,r}}
{\sum_{j=1}^{R_p}A_{c,j}}.
\label{eq:aligned_aggregation}
\end{align}

\noindent Equation~\eqref{eq:aligned_aggregation} is part of the proposed DP-PMA method. It combines the aligned client means, within-component covariances, between-client variation, and component masses to form the global private distribution.

\begin{equation}
{G}_{c}
=
\sum_{r=1}^{R_p}
\widetilde{\pi}_{c,r}
\mathcal{N}
\left(
\widetilde{\boldsymbol{\mu}}_{c,r},
\operatorname{diag}
\left(
\widetilde{\boldsymbol{\Sigma}}_{c,r}
\right)
\right).
\label{eq:global_private_distribution}
\end{equation}

\begin{algorithm}[t]
\caption{DP-PMA}
\label{alg:dppma}
\begin{algorithmic}[1]
\Require $\{\mathbf{h}_{k,i},y_{k,i}\}_{k=1}^{K}$, $C$, $R_p$,
$(\varepsilon,\delta)$
\Ensure $\{\widetilde{\mathcal{G}}_{c}\}_{c=0}^{1}$

\For{$k=1,\ldots,K$, $c\in\{0,1\}$}
    \State \textbf{Latent clipping:}
    $\overline{\mathbf{h}}_{k,i}
    \gets
    \operatorname{Clip}(\mathbf{h}_{k,i},C)$

    \State \textbf{Fixed sharding:}
    $\{\mathcal{S}_{k,c,r}\}_{r=1}^{R_p}
    \gets
    \operatorname{Shard}(\overline{\mathbf{h}}_{k,c})$

    \State \textbf{Prototype mean:}
    $\boldsymbol{\mu}_{k,c,r}
    \gets
    \operatorname{Mean}(\mathcal{S}_{k,c,r})$

    \State \textbf{DP perturbation:}
    $\widetilde{\boldsymbol{\mu}}_{k,c,r}
    \gets
    \boldsymbol{\mu}_{k,c,r}
    +
    \mathcal{N}(\mathbf{0},\sigma_{k,c,r}^{2}\mathbf{I})$
\EndFor

\State \textbf{Component alignment:}
$\rho_{k,c}^{\star}
\gets
\operatorname{Hungarian}
\bigl(\widetilde{\boldsymbol{\mu}}_{k,c,r}\bigr)$

\State \textbf{Mixture aggregation:}
$\widetilde{\mathcal{G}}_{c}
\gets
\operatorname{Aggregate}
\bigl(
\widetilde{\boldsymbol{\mu}}_{k,c,r},
\boldsymbol{\Sigma}_{c},
\pi_{k,c,r},
\rho_{k,c}^{\star}
\bigr)$

\State \Return $\{\widetilde{\mathcal{G}}_{c}\}_{c=0}^{1}$
\end{algorithmic}
\end{algorithm}

\subsection{Private Latent Diffusion, Downstream Classification, and Privacy Scope}
\label{subsec:diffusion_classification_privacy}

\noindent For each class $c$, the server generates latent training samples from the global private mixture:

\begin{equation}
\mathbf{x}_{0}
\sim
{G}_{c},
\qquad
c\in\{0,1\}.
\label{eq:private_mixture_sampling}
\end{equation}

\noindent These mixture samples, rather than individual client latent vectors, train the class-conditional diffusion model. Following the denoising diffusion probabilistic model of Ho et al. \cite{ho2020denoising}, the forward noising process is:

\begin{equation}
\mathbf{x}_t
=
\sqrt{\overline{\alpha}_t}\mathbf{x}_0
+
\sqrt{1-\overline{\alpha}_t}
\boldsymbol{\epsilon},
\qquad
\boldsymbol{\epsilon}
\sim
\mathcal{N}(\mathbf{0},\mathbf{I}),
\qquad
\overline{\alpha}_t
=
\prod_{s=1}^{t}(1-\beta_s).
\label{eq:forward_diffusion}
\end{equation}

\noindent The denoiser is trained with the DDPM noise-prediction objective \cite{ho2020denoising}:

\begin{equation}
\mathcal{L}_{\mathrm{diff}}
=
\mathbb{E}_{
\mathbf{x}_0,c,t,\boldsymbol{\epsilon}
}
\left[
\left\|
\boldsymbol{\epsilon}
-
\boldsymbol{\epsilon}_{\psi}
(\mathbf{x}_t,t,c')
\right\|_2^2
\right],
\label{eq:diffusion_loss}
\end{equation}

\noindent where $\boldsymbol{\epsilon}_{\psi}$ is the conditional denoising network and $c'$ is replaced by an unconditional class token during a fraction of the training iterations. At generation time, conditional and unconditional predictions are combined through classifier-free guidance \cite{ho2022classifierfree}:

\begin{equation}
\widehat{\boldsymbol{\epsilon}}_t
=
\boldsymbol{\epsilon}_{u}
+
s
\left(
\boldsymbol{\epsilon}_{c}
-
\boldsymbol{\epsilon}_{u}
\right),
\qquad
s=1.5.
\label{eq:classifier_free_guidance}
\end{equation}

\noindent Here, $\boldsymbol{\epsilon}_{u}$ and $\boldsymbol{\epsilon}_{c}$ are the unconditional and class-conditional
noise estimates, respectively, and $s$ is the guidance scale. Starting from $\mathbf{x}_{T_d}\sim\mathcal{N}(\mathbf{0},\mathbf{I})$, we use the deterministic DDIM update of Song et al. \cite{song2021ddim}. The estimated clean representation and reverse update are:

\begin{equation}
\widehat{\mathbf{x}}_{0}^{(t)}
=
\frac{
\mathbf{x}_t
-
\sqrt{1-\overline{\alpha}_t}
\widehat{\boldsymbol{\epsilon}}_t
}{
\sqrt{\overline{\alpha}_t}
},
\qquad
\mathbf{x}_{t-1}
=
\sqrt{\overline{\alpha}_{t-1}}
\widehat{\mathbf{x}}_{0}^{(t)}
+
\sqrt{1-\overline{\alpha}_{t-1}}
\widehat{\boldsymbol{\epsilon}}_t.
\label{eq:reverse_diffusion}
\end{equation}

\noindent After the final reverse step, the possible representation is:

\begin{equation}
\mathbf{z}_{\mathrm{syn}}
=
\mathbf{x}_{0}^{\mathrm{gen}}
\in\mathbb{R}^{64}.
\label{eq:synthetic_latent}
\end{equation}

\noindent The synthetic latent dataset is written as:

\begin{equation}
\mathcal{D}_{\mathrm{syn}}
=
\left\{
\left(
\mathbf{z}_{\mathrm{syn},i},
y_i
\right)
\right\}_{i=1}^{N_{\mathrm{syn}}}.
\label{eq:synthetic_dataset}
\end{equation}

\noindent A classifier with architecture $64\!\rightarrow\!192\!\rightarrow\!96\!\rightarrow\!1$ is trained on $\mathcal{D}_{\mathrm{syn}}$. Its probability and binary cross-entropy objective are:

\begin{equation}
\widehat{p}_i
=
\sigma
\left(
g_{\boldsymbol{\omega}}
(\mathbf{z}_{\mathrm{syn},i})
\right),
\qquad
\mathcal{L}_{\mathrm{syn}}
=
-\frac{1}{N_{\mathrm{syn}}}
\sum_{i=1}^{N_{\mathrm{syn}}}
\left[
y_i\log\widehat{p}_i
+
(1-y_i)
\log(1-\widehat{p}_i)
\right].
\label{eq:synthetic_classifier}
\end{equation}

\noindent The trained classifier is evaluated on real latent representations through the train-synthetic-test-real protocol. The formal $(\varepsilon,\delta)$ applies to the released clipped shard means in Eq.~\eqref{eq:private_prototype} under fixed-size, label-preserving replace-one adjacency. By the post-processing of differential privacy \cite{dwork2006calibrating}, subsequent alignment, mixture construction, diffusion training, and latent generation consume no additional privacy budget with respect to the protected prototype release. FedDP-PALD generates 64-dimensional latent representations, not raw chest X-ray images or ECG waveforms.

\section{Experimental Setup}
\label{sec:experimental_setup}

\noindent All experiments were implemented in Python using PyTorch, scikit-learn and executed in a Kaggle environment equipped with two NVIDIA T4 GPUs. Reproducibility was maintained by fixing the random states of Python, NumPy, PyTorch, and CUDA, while deterministic cuDNN operations were enabled. The principal classification experiments were repeated using three random seeds. Model checkpoints were selected using validation AUROC, with validation loss applied as a secondary criterion. Classification thresholds were determined only from the validation data and subsequently applied to the test data without further adjustment. Results from repeated experiments are reported as the mean and standard deviation.

\noindent Classification performance was assessed from the confusion matrix entries: true negatives, false positives, false negatives, and true positives together with accuracy, precision, recall, specificity, negative predictive value, and F1 score \cite{tharwat2021classification}. Ranking performance was measured using the area under the receiver operating characteristic curve (AUROC) \cite{fawcett2006roc} and the area under the precision-recall curve (AUPRC) \cite{saito2015precision}. Balanced accuracy \cite{brodersen2010balanced} and the Matthews correlation coefficient (MCC) \cite{matthews1975comparison} were included to account for class imbalance, whereas the Brier score and 15-bin expected calibration error (ECE) assessed probabilistic calibration \cite{brier1950verification,guo2017calibration}. Privacy was evaluated through repeated membership-inference attack AUROC and the separation between member and non-member scores \cite{shokri2017membership}. Synthetic-data fidelity was examined using nearest-neighbour distances, the near-duplicate fraction, class-conditional Fr\'echet latent distance, maximum mean discrepancy, synthetic precision, recall, density, and coverage \cite{heusel2017gans,gretton2012kernel,naeem2020reliable}. Membership-inference experiments were repeated 100 times, while paired comparisons used 5000 bootstrap resamples to obtain 95\% percentile intervals.

\section{Results and Discussion}
\label{sec:results_discussion}

\begin{figure*}[!htbp]
    \centering
    \includegraphics[
        width=\textwidth,
        height=0.70\textheight,
        keepaspectratio
    ]{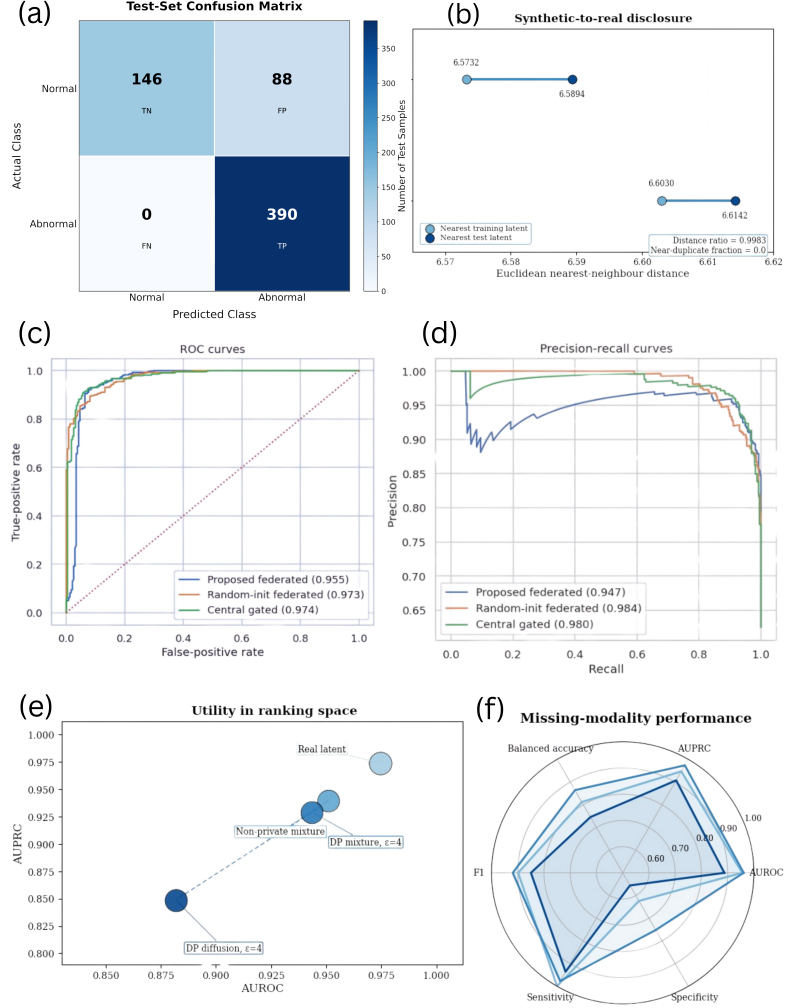}
    \caption{Joint performance, privacy, utility, and missing modality analysis of FedDP-PALD. (a) Representative test confusion matrix for the downstream classifier trained on DP latent-diffusion samples at $\varepsilon=4$ and classifier seed 42. (b) Euclidean synthetic-to-real nearest-neighbour distances against the training and held-out real latent sets. (c) Test ROC curves for the proposed pretrained federated model, the random-initialized federated model, and the centralized gated model. (d) Corresponding precision-recall curves. (e) Three-seed mean AUROC and AUPRC for the real-latent reference, non-private prototype mixture, DP prototype mixture, and DP latent diffusion. (f) Performance under complete, image-only, and ECG-only modality conditions.}
    \label{fig:combined_results}
\end{figure*}

\noindent We examine FedDP-PALD four related perspectives: multimodal diagnostic discrimination, resistance to membership inference, utility of private synthetic representations, and robustness to missing modalities. In Figure~\ref{fig:combined_results}, we present the principal findings.

\noindent FedDP-PALD substantially weakens summary-level membership inference while preserving high sensitivity and F1 score in the train-synthetic-test-real experiment. The remaining discrepancy is concentrated in rank ordering, specificity, calibration, and class-conditional distributional coverage. These distinctions are clinically relevant because a model may retain an apparently strong thresholded F1 score while still assigning poorly separated probability scores to difficult negative observations.

\subsection{Federated Multimodal Representation Performance}
\label{subsec:federated_results}

\noindent The primary pretrained federated model reached a validation AUROC of $0.9954$, an AUPRC of $0.9984$, an F1 score of $0.9805$, and a balanced accuracy of $0.9735$. On the independent test set, its AUROC was $0.9552$, with an AUPRC of $0.9469$, accuracy of $0.8606$, F1 score of $0.8997$, and balanced accuracy of $0.8141$. The validation-selected threshold was $0.7359$. At this threshold, the model detected all 390 positive test observations, resulting in sensitivity and negative predictive value of $1.0000$. However, 87 of the 234 negative
observations were predicted as abnormal. Consequently, specificity was $0.6282$, precision was $0.8176$, and the Matthews correlation coefficient was $0.7167$. The test Brier score and expected calibration error were $0.1415$ and $0.1629$, respectively. The substantial difference between validation and test loss ($0.0886$ versus $0.7496$) indicates distributional change between the two splits, despite the continued high test AUROC. This pattern is consistent with a model that preserves broad discrimination but becomes overconfident or less well calibrated on the independent test distribution. Temperature scaling reduced the Brier score from $0.1415$ to $0.1378$, but did not improve the expected calibration error, which changed slightly from $0.1629$ to $0.1634$. Thus, a single temperature parameter was insufficient to correct the full distributional calibration shift.

Table~\ref{tab:encoder_comparison} compares the proposed model with
centralized, unimodal, and initialization baselines.

\begin{table*}[t]
\centering
\caption{Test performance of the federated, centralized, and unimodal representation-learning models. Values for the first seven models are from seed 42.}
\label{tab:encoder_comparison}
\small
\resizebox{\textwidth}{!}{%
\begin{tabular}{lccccccc}
\toprule
\textbf{Model}
& \textbf{AUROC}
& \textbf{AUPRC}
& \textbf{Bal. Acc.}
& \textbf{Sensitivity}
& \textbf{Specificity}
& \textbf{F1}
& \textbf{Brier} \\
\midrule
FedAvg gated attention, pretrained
& 0.9552 & 0.9469 & 0.8141 & \textbf{1.0000} & 0.6282 & 0.8997 & 0.1415 \\

FedAvg gated attention, random initialization
& 0.9735 & \textbf{0.9842} & 0.8209 & 0.9923 & 0.6496 & 0.9010 & 0.1339 \\

Central gated attention, pretrained
& \textbf{0.9736} & 0.9796 & 0.8462 & 0.9872 & 0.7051 & 0.9123 & 0.0904 \\

Central concatenation, pretrained
& 0.9721 & 0.9816 & 0.8359 & 0.9923 & 0.6795 & 0.9085 & \textbf{0.0812} \\

Central mean fusion, pretrained
& 0.9637 & 0.9768 & \textbf{0.8675} & 0.9744 & \textbf{0.7607} & \textbf{0.9201} & 0.0864 \\

Image only, pretrained
& 0.9595 & 0.9738 & 0.8308 & 0.9821 & 0.6795 & 0.9033 & 0.1370 \\

ECG only
& 0.8315 & 0.9029 & 0.7218 & 0.7000 & 0.7436 & 0.7552 & 0.2470 \\
\bottomrule
\end{tabular}}
\end{table*}

\noindent The centralized gated model produced a higher AUROC than the proposed federated model, $0.9736$ versus $0.9552$, and its specificity increased from $0.6282$ to $0.7051$. Central mean fusion produced the highest balanced accuracy and F1 score among these seed-42 comparisons, $0.8675$ and $0.9201$, respectively. Therefore, the results do not support a claim that federated training intrinsically outperforms centralized optimization. Its contribution is instead the ability to learn without transferring the original local X-ray and ECG data. The random-initialized FedAvg model also exceeded the pretrained federated model in AUROC and AUPRC for seed 42. A paired bootstrap comparison estimated the pretrained-minus-random-initialized AUROC difference as $-0.0182$, with a 95\% interval of $[-0.0393,-0.0006]$. The corresponding AUPRC difference was $-0.0373$, with a 95\% interval of $[-0.0683,-0.0116]$. Differences in balanced accuracy and F1 were much smaller: $-0.0017$ and $+0.0005$, respectively, and both intervals included zero. Pretraining therefore did not provide a consistent ranking advantage in the present pseudo-paired modality. Across the three primary federated seeds, AUROC values were $0.9552$, $0.9659$, and $0.9155$, while AUPRC values were $0.9469$, $0.9735$, and $0.9433$. The corresponding F1 scores were $0.8997$, $0.9013$, and $0.9005$. Hence, thresholded F1 was highly stable, whereas AUROC was more sensitive to initialization. This difference recurs in the synthetic-latent experiments and shows that one threshold-based statistic cannot fully describe model stability.

\subsection{Downstream Utility of Private Synthetic Latents}
\label{subsec:synthetic_utility}

\noindent The primary selected $\varepsilon$ experiment trained the diffusion model from 10,000 samples drawn from the released DP mixture and possible a balanced synthetic dataset containing 2,500 observations per class. Across three downstream-classifier seeds, training on these synthetic latents and testing on real test latents produced an AUROC of $0.9057\pm0.0503$, an AUPRC of $0.8747\pm0.0830$, an F1 score of $0.8993\pm0.0006$, balanced accuracy of $0.8134\pm0.0012$, sensitivity of $1.0000\pm0.0000$, and specificity of $0.6268\pm0.0025$. The real-latent ceiling produced an AUROC of $0.9747\pm0.0132$, AUPRC of $0.9741\pm0.0261$, F1 score of $0.9026\pm0.0026$, balanced accuracy of $0.8211\pm0.0061$, sensitivity of $0.9983\pm0.0015$, and specificity of $0.6439\pm0.0137$. Thus, the primary synthetic AUROC was lower by approximately $0.0690$, while the F1 difference was only $0.0033$. This contrast indicates that the synthetic classifier retained the decision boundary required for high sensitivity, but did not reproduce the same fine-grained ranking of observations as the real-latent classifier.

\noindent A bootstrap analysis of the representative seed-42 classifiers reinforced this distinction. The DP diffusion-minus-real AUROC difference was $-0.1098$, with a 95\% interval of $[-0.1506,-0.0756]$, and the AUPRC difference was $-0.1760$, with a 95\% interval of $[-0.2238,-0.1283]$. By comparison, the balanced accuracy difference was $-0.0021$, with an interval of $[-0.0182,0.0137]$, and the F1 difference was $-0.0010$, with an interval of $[-0.0086,0.0065]$. The ranking loss was therefore material, whereas the thresholded classification differences were not clearly distinguishable in this bootstrap experiment. The later standardized generator comparison used a common diffusion seed, a fixed synthesis size, three classifier seeds, and an identical evaluation procedure for all generators. These results are reported in Table~\ref{tab:generator_comparison} and are also represented in Figure~\ref{fig:combined_results}(e).

\begin{table*}[t]
\centering
\caption{Three-seed downstream utility under the standardized train-synthetic-test-real protocol.}
\label{tab:generator_comparison}
\small
\resizebox{\textwidth}{!}{%
\begin{tabular}{lcccccc}
\toprule
\textbf{Training representation}
& \textbf{AUROC}
& \textbf{AUPRC}
& \textbf{Bal. Acc.}
& \textbf{Sensitivity}
& \textbf{Specificity}
& \textbf{F1} \\
\midrule
Real latent, size-matched
& $\mathbf{0.9829\pm0.0005}$
& $\mathbf{0.9893\pm0.0004}$
& $\mathbf{0.8303\pm0.0015}$
& $0.9983\pm0.0015$
& $\mathbf{0.6624\pm0.0043}$
& $\mathbf{0.9072\pm0.0006}$ \\

Real latent, full ceiling
& $0.9747\pm0.0132$
& $0.9741\pm0.0261$
& $0.8211\pm0.0061$
& $0.9983\pm0.0015$
& $0.6439\pm0.0137$
& $0.9026\pm0.0026$ \\

Non-private prototype mixture
& $0.9509\pm0.0250$
& $0.9395\pm0.0417$
& $0.7949\pm0.0000$
& $\mathbf{1.0000\pm0.0000}$
& $0.5897\pm0.0000$
& $0.8904\pm0.0000$ \\

DP prototype mixture, $\varepsilon=4$
& $0.9435\pm0.0339$
& $0.9291\pm0.0533$
& $0.8013\pm0.0111$
& $\mathbf{1.0000\pm0.0000}$
& $0.6026\pm0.0222$
& $0.8935\pm0.0053$ \\

DP latent diffusion, $\varepsilon=4$
& $0.8817\pm0.0737$
& $0.8487\pm0.1079$
& $0.8134\pm0.0012$
& $\mathbf{1.0000\pm0.0000}$
& $0.6268\pm0.0025$
& $0.8993\pm0.0006$ \\
\bottomrule
\end{tabular}}
\end{table*}

\noindent The prototype mixtures retained more ranking information than the diffusion outputs. At $\varepsilon=4$, direct samples from the DP prototype mixture achieved an AUROC of $0.9435$, compared with $0.8817$ after diffusion. Nevertheless, diffusion increased specificity from $0.6026$ to $0.6268$ and raised the F1 score from $0.8935$ to $0.8993$. This result suggests that the diffusion model changed the geometry of the possible distribution in a way that improved the validation-selected operating point while weakening global score ordering. The primary and standardized $\varepsilon=4$ diffusion experiments produced AUROCs of $0.9057\pm0.0503$ and $0.8817\pm0.0737$, respectively. The difference arises because the primary branch and standardized rerun used separately selected diffusion checkpoints and randomization protocols. The primary value is retained for consistency with the headline experiment, whereas Table~\ref{tab:generator_comparison} and Figure~\ref{fig:combined_results}(e) use the standardized comparison. These values should not be combined as though they arose from one identical repeated experiment. The representative confusion matrix in Figure~\ref{fig:combined_results}(a) contains 146 true negatives, 88 false positives, no false negatives, and 390 true positives. This gives accuracy $0.8590$, precision $0.8159$, sensitivity $1.0000$, specificity $0.6239$, F1 $0.8986$, and balanced accuracy $0.8120$. The absence of false negatives is encouraging for a screening-oriented application. However, the 88 false-positive predictions indicate that such a classifier would require confirmatory review before clinical use.

\subsection{Privacy-Budget Analysis}
\label{subsec:privacy_results}

\noindent The repeated released-summary attack provides the evidence of the effect of DP-PMA. Without differential privacy, the attack AUROC was $0.6229\pm0.0026$. Applying differential privacy reduced it to $0.5093\pm0.0026$ at $\varepsilon=8$, $0.5068\pm0.0024$ at $\varepsilon=4$, $0.5029\pm0.0023$ at $\varepsilon=2$, and $0.5016\pm0.0023$ at $\varepsilon=1$. Values near 0.5 indicate that the attack could not reliably distinguish members from non-members using the released prototype summaries. The mean member-non-member score gap also decreased steadily from $82.4251$ without privacy to $47.0571$, $30.3635$, $15.1606$, and $8.1640$ for $\varepsilon=8$, $4$, $2$, and $1$, respectively. The paired attack-AUROC reductions relative to the non-private release were $-0.1136$, $-0.1161$, $-0.1200$, and $-0.1213$. Every one of the 100 paired repetitions produced a lower attack AUROC for each DP budget than for the corresponding non-private release.

\begin{table*}[t]
\centering
\caption{Standardized privacy-budget results. The attack was repeated 100 times per modality, while downstream utility was evaluated across three classifier seeds.}
\label{tab:privacy_sweep}
\small
\resizebox{\textwidth}{!}{%
\begin{tabular}{lcccccc}
\toprule
\textbf{Release}
& \textbf{Attack AUROC}
& \textbf{95\% attack range}
& \textbf{Synthetic AUROC}
& \textbf{Synthetic AUPRC}
& \textbf{Specificity}
& \textbf{F1} \\
\midrule
Non-private
& $0.6229\pm0.0026$
& $[0.6171,0.6272]$
& $0.8615\pm0.0754$
& $0.8224\pm0.1024$
& $0.6254\pm0.0025$
& $0.8990\pm0.0006$ \\

$\varepsilon=8$
& $0.5093\pm0.0026$
& $[0.5039,0.5133]$
& $0.8732\pm0.0800$
& $0.8391\pm0.1169$
& $0.6254\pm0.0025$
& $0.8990\pm0.0006$ \\

$\varepsilon=4$
& $0.5068\pm0.0024$
& $[0.5020,0.5105]$
& $0.8817\pm0.0737$
& $0.8487\pm0.1079$
& $\mathbf{0.6268\pm0.0025}$
& $\mathbf{0.8993\pm0.0006}$ \\

$\varepsilon=2$
& $0.5029\pm0.0023$
& $[0.4986,0.5066]$
& $0.9057\pm0.0576$
& $0.8785\pm0.0857$
& $\mathbf{0.6268\pm0.0025}$
& $\mathbf{0.8993\pm0.0006}$ \\

$\varepsilon=1$
& $\mathbf{0.5016\pm0.0023}$
& $\mathbf{[0.4968,0.5055]}$
& $\mathbf{0.9059\pm0.0535}$
& $\mathbf{0.8826\pm0.0777}$
& $\mathbf{0.6268\pm0.0025}$
& $\mathbf{0.8993\pm0.0006}$ \\
\bottomrule
\end{tabular}}
\end{table*}

\noindent The downstream results deviated from a conventional monotonic privacy-utility curve. Stronger privacy was not associated with a reduction in mean AUROC. Instead, the highest standardized values occurred at $\varepsilon=1$ and $\varepsilon=2$. This pattern should not be interpreted as evidence that increasing DP noise generally improves clinical utility. Each privacy budget produced a newly perturbed mixture and a separately selected diffusion checkpoint. The injected noise may have acted as a regularizer, while checkpoint selection and the limited number of classifier seeds may also have contributed to the observed variability. The stable F1 score across privacy budgets mainly reflects the common validation-selected operating point and the positive-class dominance of the test data. The privacy guarantee must also be interpreted at the correct scope. It applies to the clipped, perturbed fixed-shard prototype means under the specified replace-one adjacency. The FedAvg model parameters, encoder gradients, public class counts, validation-derived clipping radius, covariance estimates, and validation latents were not covered by this formal guarantee. Accordingly, the reported attack reduction supports the privacy of the released summary mechanism, rather than a claim of end-to-end patient-level differential privacy. Additional attacks against the unprotected federated encoder produced AUROC values of $0.5849$ for the loss-based attack and $0.5539$ for both confidence- and entropy-based attacks. These values are above chance, although substantially below perfect inference. They confirm that local data retention alone does not make the preceding federated training phase private.

\subsection{Nearest-Neighbour Disclosure and Duplicate Analysis}
\label{subsec:disclosure_results}

\noindent Figure~\ref{fig:combined_results}(b) compares every possible representation with its closest real training and held-out test latent. The median nearest-neighbour distances were $6.5732$ to the training set and $6.5894$ to the held-out set. The corresponding mean distances were $6.6030$ and $6.6142$. Their ratio was $0.9983$, and the near-duplicate fraction was $0.0000$. The nearly equal train and held-out distances provide no evidence that the generator was systematically closer to the representations used to construct the released training summary. Furthermore, no possible sample crossed the near-duplicate criterion. This finding complements the membership attack, but it should not be treated as a formal privacy proof. Nearest-neighbour distance is sensitive to latent scaling, dimension, and the selected duplicate threshold. It measures local geometric disclosure rather than the worst-case influence controlled by differential privacy.

\subsection{Distributional Fidelity and Coverage}
\label{subsec:fidelity_results}

\noindent The fidelity analysis shows a more qualified result than the classification metrics alone. Table~\ref{tab:fidelity_results} reports class-conditional Fréchet latent distance, radial-basis-function MMD, and real-sample recall. Lower Fréchet distance and MMD indicate closer distributional agreement, whereas higher recall indicates broader coverage of the real test distribution.

\begin{table*}[t]
\centering
\caption{Class-conditional fidelity of the standardized synthetic
representations.}
\label{tab:fidelity_results}
\small
\resizebox{\textwidth}{!}{%
\begin{tabular}{llccc}
\toprule
\textbf{Generator}
& \textbf{Class}
& \textbf{Fréchet latent distance}
& \textbf{MMD}
& \textbf{Recall} \\
\midrule
Non-private prototype mixture
& Negative & 88.3108 & 0.2816 & 0.3761 \\
Non-private prototype mixture
& Positive & \textbf{10.1404} & \textbf{0.0878} & 0.9718 \\

DP prototype mixture, $\varepsilon=4$
& Negative & 106.9650 & 0.2202 & 0.5214 \\
DP prototype mixture, $\varepsilon=4$
& Positive & 15.6314 & 0.0924 & \textbf{0.9897} \\

DP latent diffusion, $\varepsilon=4$
& Negative & \textbf{76.8075} & \textbf{0.0720} & \textbf{0.7650} \\
DP latent diffusion, $\varepsilon=4$
& Positive & 93.9246 & 0.5006 & 0.1282 \\
\bottomrule
\end{tabular}}
\end{table*}

\noindent The DP diffusion model represented the negative class more broadly than either prototype sampler: negative-class recall increased to $0.7650$, while MMD decreased to $0.0720$. The positive class exhibited the opposite pattern. Positive-class recall declined to $0.1282$, whereas the Fréchet distance and MMD rose to $93.9246$ and $0.5006$, respectively. These findings suggest that the diffusion model distributed probability mass more widely across the negative latent region while substantially contracting the support of the positive class. Synthetic precision, density, and coverage were zero for every generator under the selected five-neighbour PRDC criterion. Such values should not be interpreted as evidence that the synthetic data lacked predictive information, since the train-synthetic-test-real classifiers still achieved high F1 scores and moderate-to-high AUROC values. Rather, they indicate insufficient overlap between the strict local neighbourhoods of the synthetic and real test distributions under the current latent scaling. The coexistence of useful classification performance and weak PRDC scores therefore illustrates why downstream accuracy alone provides an incomplete assessment of synthetic medical data quality. An earlier direct real-versus-synthetic diagnostic produced latent-mean gaps of $5.5069$ for the negative class and $6.2393$ for the positive class. The real latent norms were $7.4149$ and $10.1472$, compared with synthetic norms of $6.9446$ and $7.3602$. The lower positive synthetic norm is consistent with contraction of the positive distribution toward a narrower region.

\subsection{Ranking Errors and Score Compression}
\label{subsec:ranking_failure}

\noindent The failure analysis demonstrates why F1 remained high while AUROC and AUPRC declined. Among the true-positive samples, the synthetic classifier concentrated most predicted probabilities near $0.9328$. Of the 390 positive test samples, 294 received a probability that rounded to $0.9328$, while a further 45 rounded to $0.9329$. The positive-class 90th-minus-10th-percentile score range was only $0.0001$, compared with $0.0002$ for the highly confident real-latent classifier. More importantly, the synthetic positive logits had a mean of $2.6294$ and a standard deviation of only $0.0085$, whereas the real-latent classifier produced a mean logit of $6.8707$ with a standard deviation of $0.7352$. The negative class retained a wider synthetic score range, although difficult negative samples were frequently shifted toward higher predicted probabilities. Twenty-nine hard negatives were identified from the largest synthetic-over-real probability differences. Their mean synthetic probability was $0.8172$, compared with a mean real-classifier probability of $0.3339$. The resulting average synthetic-minus-real gap was $0.4833$. The synthetic scores were also strongly associated with local latent geometry: the correlation with the positive-minus-negative centroid margin was $-0.9231$, while the correlation with the fraction of positive observations among the 25 nearest neighbours was $0.9448$. For the 28 borderline negative samples whose positive-neighbour fractions lay between $0.2$ and $0.8$, the synthetic classifier assigned a mean probability of $0.6691$, compared with $0.3746$ for the real-latent classifier.

\subsection{Missing-Modality Performance}
\label{subsec:missing_modality_results}

Figure~\ref{fig:combined_results}(f) and Table~\ref{tab:missing_modality} examine the same federated encoder when both modalities, only the image, or only the ECG are supplied.

\begin{table*}[t]
\centering
\caption{Test performance under complete and missing-modality conditions. Thresholds were selected independently from the corresponding validation condition.}
\label{tab:missing_modality}
\small
\resizebox{\textwidth}{!}{%
\begin{tabular}{lcccccccccc}
\toprule
\textbf{Condition}
& \textbf{AUROC}
& \textbf{AUPRC}
& \textbf{Accuracy}
& \textbf{Precision}
& \textbf{Sensitivity}
& \textbf{Specificity}
& \textbf{NPV}
& \textbf{F1}
& \textbf{Bal. Acc.}
& \textbf{MCC} \\
\midrule
Both modalities
& 0.9552 & 0.9469 & 0.8590 & 0.8159 & 1.0000
& 0.6239 & 1.0000 & 0.8986 & 0.8120 & 0.7135 \\

Image only
& 0.9613 & 0.9737 & 0.8926 & 0.8679 & 0.9769
& 0.7521 & 0.9514 & 0.9192 & 0.8645 & 0.7728 \\

ECG only
& 0.8874 & 0.9074 & 0.7933 & 0.7783 & 0.9359
& 0.5556 & 0.8387 & 0.8498 & 0.7457 & 0.5506 \\
\bottomrule
\end{tabular}}
\end{table*}

\noindent The model preserved predictive capability when either modality was unavailable, supporting the effectiveness of the availability masks and modality-dropout strategy. Under the ECG-only modality, the model achieved an AUROC of $0.8874$, an AUPRC of $0.9074$, a sensitivity of $0.9359$, and an F1 score of $0.8498$. The image-only modality produced stronger results, with an AUROC of $0.9613$, an AUPRC of $0.9737$, a balanced accuracy of $0.8645$, and an F1 score of $0.9192$. The image-only modality also surpassed the two-modality modality in balanced accuracy, specificity, F1 score, and AUPRC. Paired bootstrap analysis yielded a both-minus-image balanced-accuracy difference of $-0.0526$, with a 95\% confidence interval of $[-0.0795,-0.0256]$. The corresponding F1 difference was $-0.0206$, with a confidence interval of $[-0.0350,-0.0054]$. Incorporating both modalities increased sensitivity by $0.0231$, whereas specificity decreased by $0.1282$. The AUROC difference was $-0.0060$, with a confidence interval of $[-0.0263,0.0134]$; the inclusion of zero in this interval indicates limited evidence of an AUROC difference after multiplicity adjustment. The two-modality modality clearly outperformed the ECG-only modality. The paired AUROC difference was $0.0678$, with a 95\% confidence interval of $[0.0366,0.1009]$. Balanced accuracy, F1 score, and sensitivity increased by $0.0662$, $0.0488$, and $0.0641$, respectively. The image-only modality also surpassed the ECG-only modality, with gains of $0.0739$ in AUROC, $0.0663$ in AUPRC, $0.1188$ in balanced accuracy, $0.0694$ in F1 score, and $0.1966$ in specificity. The stronger image-only performance should be considered in relation to the data-construction strategy. The X-ray images and ECG segments were paired according to class labels rather than patient identity. Consequently, the ECG branch contributed class-consistent physiological information without representing the same clinical subject as the corresponding image. Multimodal fusion may therefore increase sensitivity by introducing an additional positive-class signal while also increasing false-positive predictions. Overall, the findings demonstrate resilience to missing modalities, although the current evidence provides insufficient support for superiority of pseudo-paired multimodal fusion over the image-only modality.

\subsection{Gate and Attention Behaviour}
\label{subsec:gate_attention}

\noindent The learned gates differed strongly by class. For negative observations, the mean image and ECG gates were $0.1094$ and $0.0444$. For positive observations, they increased to $0.7213$ and $0.0903$. Thus, the model placed substantially greater gated weight on image information for the positive class, while the ECG gate remained comparatively small. The attention matrices showed a related pattern. For positive samples, the image token assigned a mean weight of $0.8809$ to itself and $0.1191$ to the ECG token. The ECG token assigned $0.9533$ of its attention to the image token and only $0.0467$ to itself. For negative samples, the corresponding values were less concentrated: image-to-image attention was $0.6273$, image-to-ECG attention was $0.3727$, ECG-to-image attention was $0.6871$, and ECG-to-ECG attention was $0.3129$. The attention mechanism therefore learned an image dominant positive-class representation, consistent with the stronger image-only results.

\subsection{Ablation Studies}
\label{subsec:architecture_ablation}

The results of the three-seed ablation study are presented in Table~\ref{tab:architecture_ablation}. The full FedDP-PALD architecture achieved the most stable F1 score of $0.9005\pm0.0008$, together with a balanced accuracy of $0.8251\pm0.0141$ and a specificity of $0.6681\pm0.0550$. Removing the token gates reduced the F1 score to $0.8920\pm0.0140$, balanced accuracy to $0.8011\pm0.0263$, and specificity to $0.6083\pm0.0477$, demonstrating the contribution of the gating mechanism to reliable modality weighting and class-level decision performance. Excluding ChestMNIST pretraining produced the lowest balanced accuracy of $0.7950\pm0.0225$ and the lowest F1 score of $0.8888\pm0.0091$, highlighting the value of transferred image representations for the selected classification operating point. Removing both attention and token gating increased the mean AUROC and AUPRC to $0.9696\pm0.0079$ and $0.9773\pm0.0095$, respectively, while the F1 score remained comparable at $0.9014\pm0.0156$. Excluding modality dropout achieved the highest mean AUROC of $0.971 \pm0.0115$, the highest AUPRC of $0.9811\pm0.0083$, and the lowest Brier score of $0.1133\pm0.0175$. However, this configuration also showed greater variation in specificity and F1 score across the three seeds. These findings indicate that simpler configurations can provide stronger ranking or calibration performance, whereas the full architecture offers greater stability in threshold-based classification. Overall, the ablation results demonstrate that token gating and encoder pretraining contribute to balanced and consistent decision performance, while attention and modality dropout influence the trade-off among ranking ability, calibration, and seed-level stability.

\begin{table*}[t]
\centering
\caption{Performance metrics for the ablation studies comprising different combinations.}
\label{tab:architecture_ablation}
\small
\resizebox{\textwidth}{!}{%
\begin{tabular}{lcccccc}
\toprule
\textbf{Configuration}
& \textbf{AUROC}
& \textbf{AUPRC}
& \textbf{Bal. Acc.}
& \textbf{Specificity}
& \textbf{F1}
& \textbf{Brier} \\
\midrule
Full proposed architecture
& $0.9455\pm0.0266$
& $0.9546\pm0.0165$
& $\mathbf{0.8251\pm0.0141}$
& $\mathbf{0.6681\pm0.0550}$
& $0.9005\pm0.0008$
& $0.1263\pm0.0156$ \\

Without token gates
& $0.9542\pm0.0307$
& $0.9540\pm0.0483$
& $0.8011\pm0.0263$
& $0.6083\pm0.0477$
& $0.8920\pm0.0140$
& $0.1491\pm0.0204$ \\

Without attention and token gates
& $0.9696\pm0.0079$
& $0.9773\pm0.0095$
& $0.8204\pm0.0327$
& $0.6467\pm0.0664$
& $\mathbf{0.9014\pm0.0156}$
& $0.1420\pm0.0026$ \\

Without modality dropout
& $\mathbf{0.9714\pm0.0115}$
& $\mathbf{0.9811\pm0.0083}$
& $0.8197\pm0.0522$
& $0.6496\pm0.1121$
& $0.9003\pm0.0235$
& $\mathbf{0.1133\pm0.0175}$ \\

Without ChestMNIST pretraining
& $0.9696\pm0.0097$
& $0.9808\pm0.0065$
& $0.7950\pm0.0225$
& $0.5969\pm0.0513$
& $0.8888\pm0.0091$
& $0.1236\pm0.0197$ \\
\bottomrule
\end{tabular}}
\end{table*}

\noindent The experimental findings demonstrate that FedDP-PALD learns discriminative multimodal latent representations under federated training while retaining performance when one modality is unavailable. The gated federated encoder achieved a test AUROC of $0.9552$ and an AUPRC of $0.9469$. Performance remained strong for both image-only and ECG-only inputs, although the image modality produced the best results. This pattern suggests that class-aligned pseudo-pairing provides useful multimodal information, while the X-ray representation remains the dominant source of diagnostic evidence. DP-PMA substantially reduced membership leakage from the released latent summaries. The repeated membership-inference attack AUROC decreased from $0.6229\pm0.0026$ for the non-private release to values between $0.5016$ and $0.5093$ across the examined privacy budgets. These values approach random-guessing performance and indicate that the calibrated perturbation effectively obscured membership information contained in the prototype means. The formal privacy guarantee, however, applies specifically to the released clipped prototype means and subsequent post-processing stages. The synthetic latent representations also preserved clinically relevant decision information. Classifiers trained on private diffusion-generated latents achieved a sensitivity of $1.0000$ and an F1 score near $0.8993$ when evaluated on real test latents. The lower AUROC and AUPRC relative to real-latent training indicate partial loss of fine-grained ranking information, even though the principal classification boundary remained effective. The stable F1 score across privacy budgets further suggests that FedDP-PALD maintained the selected decision threshold under increasing privacy protection. The joint privacy, utility, and fidelity analysis reveals a more detailed performance pattern. Stronger privacy consistently reduced membership-attack effectiveness, whereas downstream utility followed a non-monotonic trend across privacy budgets. Threshold-based classification remained comparatively stable, while AUROC, AUPRC, nearest-neighbour measures, and class-conditional fidelity metrics exposed variations in ranking quality and latent coverage. The main synthesis-related weakness arose from contraction of the positive-class distribution and elevated positive probabilities for negative samples located near mixed-class latent neighbourhoods. Overall, FedDP-PALD provides an effective framework for privacy-aware latent representation sharing and synthetic-data generation. DP-PMA reduces disclosure from released summaries, while the diffusion stage produces synthetic latents capable of supporting high-sensitivity downstream classification. The results also show that privacy protection, predictive utility, calibration, and distributional fidelity should be evaluated together, since strong F1 performance alone may conceal ranking errors or incomplete class coverage.

\noindent To examine sensitivity to random initialization, we further
report the performance of the full FedDP-PALD architecture for each
individual seed in Table~\ref{tab:ablation_seed_sensitivity}. The F1
score varied only from $0.8997$ to $0.9013$, with a mean of $0.9005$
and a standard deviation of $0.0008$. MCC also remained stable between
$0.7167$ and $0.7204$. Greater variation appeared in AUROC, which ranged
from $0.9155$ to $0.9659$, and in the validation-selected threshold,
which ranged from $0.3396$ to $0.7359$. These results indicate stable
threshold-based classification across seeds, together with higher
seed-level variation in probability ranking and threshold selection.

\begin{table*}[t]
\centering
\caption{Seed-sensitivity results for the full proposed FedDP-PALD
architecture.}
\label{tab:ablation_seed_sensitivity}
\small
\renewcommand{\arraystretch}{1.12}
\resizebox{\textwidth}{!}{%
\begin{tabular}{lcccccccc}
\toprule
\textbf{Seed}
& \textbf{Threshold}
& \textbf{Accuracy}
& \textbf{Precision}
& \textbf{Recall}
& \textbf{F1}
& \textbf{AUROC}
& \textbf{AUPRC}
& \textbf{MCC} \\
\midrule
42
& 0.7359
& 0.8606
& 0.8176
& 1.0000
& 0.8997
& 0.9552
& 0.9469
& 0.7167 \\

123
& 0.6284
& 0.8638
& 0.8238
& 0.9949
& 0.9013
& 0.9659
& 0.9735
& 0.7204 \\

2026
& 0.3396
& 0.8686
& 0.8548
& 0.9513
& 0.9005
& 0.9155
& 0.9433
& 0.7175 \\
\midrule
Mean
& 0.5680
& 0.8643
& 0.8321
& 0.9821
& 0.9005
& 0.9455
& 0.9546
& 0.7182 \\

Std.
& 0.2049
& 0.0040
& 0.0200
& 0.0268
& 0.0008
& 0.0266
& 0.0165
& 0.0020 \\
\bottomrule
\end{tabular}}
\end{table*}

\subsection{Limitations}
\label{subsec:limitations}

The present study has several limitations:

\begin{itemize}
    \item The X-ray images and ECG signals were paired according to class labels rather than patient identity. Consequently, the multimodal samples represent class-aligned pseudo-pairs rather than clinically matched observations from the same individual.

    \item The diffusion model generates $64$-dimensional latent representations rather than reconstructed chest X-ray images or ECG waveforms. The clinical realism of raw synthetic medical data therefore falls outside the present evaluation.

    \item The experiments used five simulated clients and three primary random seeds. A larger number of institutions, communication rounds, and repeated runs would provide a stronger assessment of stability under federated heterogeneity.

    \item Positive-class fidelity decreased after diffusion, as reflected by reduced class-conditional recall and increased Fréchet latent distance and MMD. This behaviour indicates incomplete coverage of the positive latent distribution.

    \item The image modality contributed more strongly than the ECG modality, and the combined-modality input offered limited improvement over image-only inference. Evaluation on patient-matched multimodal datasets is required to determine the full benefit of the fusion architecture.

    \item Membership inference, nearest-neighbour disclosure, and duplicate analysis represent selected privacy attacks. Additional attacks involving gradients, reconstruction, attribute inference, and model inversion would provide a broader security assessment.

    \item Thresholds and checkpoints were selected using validation performance. Evaluation across external hospitals and independent acquisition systems would provide stronger evidence of generalizability and calibration.
\end{itemize}

\section{Conclusion}
\label{sec:conclusion}

\noindent We present FedDP-PALD, a privacy-preserving federated latent diffusion framework for multimodal medical data synthesis. The proposed model combines gated image-ECG fusion, modality masks, DP prototype-mixture aggregation, and class-conditional latent diffusion. The federated encoder achieved a test AUROC of $0.9552$ and an AUPRC of $0.9469$, while retaining useful performance under image-only and ECG-only inputs. DP-PMA reduced membership-inference attack AUROC from $0.6229\pm0.0026$ to values between $0.5016$ and $0.5093$. Classifiers trained on private synthetic latents achieved a sensitivity of $1.0000$ and an F1 score near $0.8993$. These findings show that FedDP-PALD can protect released latent summaries while preserving substantial downstream utility. Future work will extend privacy protection to federated model updates, use patient-matched multimodal data, improve class-conditional coverage, and validate the framework across larger clinical federations.



\section*{Declaration of Competing Interest}
The authors declare that they have no known competing financial interests or personal relationships that could have appeared to influence the work reported in this paper.

\section*{Acknowledgements}
Collate acknowledgements in a separate section at the end of the article before the references and do not, therefore, include them on the title page, as a footnote to the title or otherwise. List here those individuals who provided help during the research (e.g., providing language help, writing assistance or proof reading the article, etc.).

\section*{Data availability}
Provide a link to your data if available.

\bibliography{sample}

@article{rieke2020future,
  title   = {The Future of Digital Health with Federated Learning},
  author  = {Rieke, N. and others},
  journal = {npj Digital Medicine},
  volume  = {3},
  pages   = {119},
  year    = {2020},
  doi     = {10.1038/s41746-020-00323-1}
}

@article{sheller2020federated,
  title   = {Federated Learning in Medicine: Facilitating Multi-Institutional
             Collaborations without Sharing Patient Data},
  author  = {Sheller, M.J. and others},
  journal = {Scientific Reports},
  volume  = {10},
  pages   = {12598},
  year    = {2020},
  doi     = {10.1038/s41598-020-69250-1}
}

@inproceedings{mcmahan2017communication,
  title     = {Communication-Efficient Learning of Deep Networks from
               Decentralized Data},
  author    = {McMahan, H.B. and Moore, E. and Ramage, D.
               and Hampson, S. and Arcas, A.B.},
  booktitle = {Proceedings of the 20th International Conference on
               Artificial Intelligence and Statistics},
  series    = {Proceedings of Machine Learning Research},
  volume    = {54},
  pages     = {1273-1282},
  publisher = {PMLR},
  year      = {2017},
  url       = {https://proceedings.mlr.press/v54/mcmahan17a.html}
}

@article{khowaja2025selffed,
  title   = {{SelfFed}: Self-Supervised Federated Learning for Data
             Heterogeneity and Label Scarcity in Medical Images},
  author  = {Khowaja, S.A. and Dev, K. and Anwar, S.M.
             and Linguraru, M.G.},
  journal = {Expert Systems with Applications},
  volume  = {261},
  pages   = {125493},
  year    = {2025},
  doi     = {10.1016/j.eswa.2024.125493}
}

@inproceedings{shokri2017membership,
  title     = {Membership Inference Attacks against Machine Learning Models},
  author    = {Shokri, R. and Stronati, M. and Song, C.
               and Shmatikov, V.},
  booktitle = {2017 IEEE Symposium on Security and Privacy},
  pages     = {3-18},
  publisher = {IEEE},
  year      = {2017},
  doi       = {10.1109/SP.2017.41}
}

@inproceedings{ye2022enhanced,
  title     = {Enhanced Membership Inference Attacks against Machine
               Learning Models},
  author    = {Ye, J. and Maddi, A. and Murakonda, S.K.
               and Bindschaedler, V. and Shokri, R.},
  booktitle = {Proceedings of the 2022 ACM SIGSAC Conference on Computer
               and Communications Security},
  pages     = {3093-3106},
  publisher = {Association for Computing Machinery},
  year      = {2022},
  doi       = {10.1145/3548606.3560675}
}

@inproceedings{dwork2006calibrating,
  title     = {Calibrating Noise to Sensitivity in Private Data Analysis},
  author    = {Dwork, C. and McSherry, F. and Nissim, K.
               and Smith, A.},
  booktitle = {Theory of Cryptography},
  series    = {Lecture Notes in Computer Science},
  volume    = {3876},
  pages     = {265-284},
  publisher = {Springer},
  year      = {2006},
  doi       = {10.1007/11681878_14}
}

@inproceedings{abadi2016deep,
  title     = {Deep Learning with Differential Privacy},
  author    = {Abadi, M. and Chu, A. and Goodfellow, I.
               and McMahan, H.B. and Mironov, I. and Talwar, K.
               and Zhang, L.},
  booktitle = {Proceedings of the 2016 ACM SIGSAC Conference on Computer
               and Communications Security},
  pages     = {308-318},
  publisher = {Association for Computing Machinery},
  year      = {2016},
  doi       = {10.1145/2976749.2978318}
}

@article{tan2022fedproto,
  title   = {{FedProto}: Federated Prototype Learning across Heterogeneous
             Clients},
  author  = {Tan, Y. and Long, G. and Liu, L. and Zhou, T.
             and Lu, Q. and Jiang, J. and Zhang, C.},
  journal = {Proceedings of the AAAI Conference on Artificial Intelligence},
  volume  = {36},
  number  = {8},
  pages   = {8432-8440},
  year    = {2022},
  doi     = {10.1609/aaai.v36i8.20819}
}

@article{ribero2022federating,
  title   = {Federating Recommendations Using Differentially Private
             Prototypes},
  author  = {Ribero, M. and Henderson, J.
             and Williamson, S. and Vikalo, H.},
  journal = {Pattern Recognition},
  volume  = {129},
  pages   = {108746},
  year    = {2022},
  doi     = {10.1016/j.patcog.2022.108746}
}

@article{wahdany2025differentially,
  title   = {Differentially Private Prototypes for Imbalanced Transfer
             Learning},
  author  = {Wahdany, D. and Jagielski, M. and Dziedzic, A.
             and Boenisch, F.},
  journal = {Proceedings of the AAAI Conference on Artificial Intelligence},
  volume  = {39},
  number  = {20},
  pages   = {20991-20999},
  year    = {2025},
  doi     = {10.1609/aaai.v39i20.35395}
}

@article{lyu2024dpldm,
  title   = {{DP-LDMs}: Differentially Private Latent Diffusion Models},
  author  = {Liu, M.F. and Lyu, S. and Vinaroz, M.
             and Park, M.},
  journal = {Transactions on Machine Learning Research},
  year    = {2024},
  url     = {https://openreview.net/forum?id=AkdQ266kHj}
}

@inproceedings{daum2024private,
  title     = {On Differentially Private 3D Medical Image Synthesis with
               Controllable Latent Diffusion Models},
  author    = {Daum, D. and Osuala, R. and Riess, A.
               and Kaissis, G. and Schnabel, J.A.
               and Folco, M.D.},
  booktitle = {Deep Generative Models: 4th MICCAI Workshop,
               DGM4MICCAI 2024},
  series    = {Lecture Notes in Computer Science},
  pages     = {139-149},
  publisher = {Springer},
  year      = {2024},
  doi       = {10.1007/978-3-031-72744-3_14}
}

@article{adams2025fidelity,
  title   = {On the Fidelity versus Privacy and Utility Trade-Off of
             Synthetic Patient Data},
  author  = {Adams, T. and Birkenbihl, C. and Otte, K.
             and Ng, H.G. and Rieling, J.A.
             and N{\"a}her, A. and Sax, U.
             and Prasser, F. and Fr{\"o}hlich, H.
             and {Alzheimer's Disease Neuroimaging Initiative}},
  journal = {iScience},
  volume  = {28},
  number  = {5},
  pages   = {112382},
  year    = {2025},
  doi     = {10.1016/j.isci.2025.112382}
}

@article{hernandez2025evaluation,
  title   = {Comprehensive Evaluation Framework for Synthetic Tabular Data
             in Health: Fidelity, Utility and Privacy Analysis of Generative
             Models with and without Privacy Guarantees},
  author  = {Hernandez, M. and Osorio-Marulanda, P.A.
             and Catalina, M. and Loinaz, L. and Epelde, G.
             and Aginako, N.},
  journal = {Frontiers in Digital Health},
  volume  = {7},
  pages   = {1576290},
  year    = {2025},
  doi     = {10.3389/fdgth.2025.1576290}
}

@misc{bao2024prototype,
  title         = {Multimodal Federated Learning with Missing Modality via
                   Prototype Mask and Contrast},
  author        = {Bao, G. and Zhang, Q. and Miao, D.
                   and Gong, Z. and Hu, L. and Liu, K.
                   and Liu, Y. and Shi, C.},
  year          = {2023},
  eprint        = {2312.13508},
  archivePrefix = {arXiv},
  primaryClass  = {cs.LG},
  doi           = {10.48550/arXiv.2312.13508}
}

@article{le2025crossmodal,
  title   = {Cross-Modal Prototype Based Multimodal Federated Learning
             under Severely Missing Modality},
  author  = {Le, H.Q. and Thwal, C.M. and Qiao, Y.
             and Tun, Y.L. and Nguyen, M.N.H. and Huh, E.
             and Hong, C.S.},
  journal = {Information Fusion},
  volume  = {122},
  pages   = {103219},
  year    = {2025},
  doi     = {10.1016/j.inffus.2025.103219}
}

@inproceedings{poudel2025feature,
  title     = {Multimodal Federated Learning with Missing Modalities
               through Feature Imputation Network},
  author    = {Poudel, P. and Chhetri, A.
               and Gyawali, P.K. and Leontidis, G.
               and Bhattarai, B.},
  booktitle = {Medical Image Understanding and Analysis},
  series    = {Lecture Notes in Computer Science},
  volume    = {15916},
  pages     = {289-299},
  publisher = {Springer},
  year      = {2025},
  doi       = {10.1007/978-3-031-98688-8_20}
}

@article{xu2025flexid,
  title   = {{FlexiD-Fuse}: Flexible Number of Inputs Multi-Modal Medical
             Image Fusion Based on Diffusion Model},
  author  = {Xu, Y. and Li, X. and Wang, Y.
             and Cheng, X. and Li, H. and Tan, H.},
  journal = {Expert Systems with Applications},
  volume  = {296},
  pages   = {128895},
  year    = {2026},
  doi     = {10.1016/j.eswa.2025.128895}
}

@article{islam2024bangla,
  title   = {Unveiling Personality Traits through Bangla Speech Using
             Morlet Wavelet Transformation and {BiG}},
  author  = {Sk., M.S.I. and Alam, M.G.R.},
  journal = {Natural Language Processing Journal},
  volume  = {9},
  pages   = {100113},
  year    = {2024},
  doi     = {10.1016/j.nlp.2024.100113}
}

@misc{islam2026vlm,
  title         = {An Explainable Vision-Language Model Framework with
                   Adaptive {PID}-Tversky Loss for Lumbar Spinal Stenosis
                   Diagnosis},
  author        = {Sk., M.S.I. and Shawon, M.M.H.
                   and Alam, M.G.R.},
  year          = {2026},
  eprint        = {2604.02502},
  archivePrefix = {arXiv},
  primaryClass  = {cs.CV},
  doi           = {10.48550/arXiv.2604.02502}
}

@misc{islam2025foundationalecg,
  title         = {FoundationalECGNet: A Lightweight Foundational Model
                   for ECG-based Multitask Cardiac Analysis},
  author        = {Islam Sk., Md. Sajeebul and Jobayer, Md.
                   and Shawon, Md. Mehedi Hasan and Alam, Md. Golam Rabiul},
  year          = {2025},
  eprint        = {2509.08961},
  archivePrefix = {arXiv},
  primaryClass  = {cs.LG},
  doi           = {10.48550/arXiv.2509.08961}
}

@article{kairouz2021advances,
  title   = {Advances and Open Problems in Federated Learning},
  author  = {Kairouz, P. and McMahan, H.B. and Avent, B.
             and others},
  journal = {Foundations and Trends in Machine Learning},
  volume  = {14},
  number  = {1-2},
  pages   = {1-210},
  year    = {2021},
  doi     = {10.1561/2200000083}
}

@inproceedings{li2020fedprox,
  title     = {Federated Optimization in Heterogeneous Networks},
  author    = {Li, T. and Sahu, A.K. and Talwalkar, A.
               and Smith, V.},
  booktitle = {Proceedings of Machine Learning and Systems},
  volume    = {2},
  pages     = {429-450},
  year      = {2020}
}

@inproceedings{balle2018improving,
  title     = {Improving the Gaussian Mechanism for Differential Privacy:
               Analytical Calibration and Optimal Denoising},
  author    = {Balle, B. and Wang, Y.},
  booktitle = {Proceedings of the 35th International Conference on
               Machine Learning},
  series    = {Proceedings of Machine Learning Research},
  volume    = {80},
  pages     = {394-403},
  publisher = {PMLR},
  year      = {2018},
  url       = {https://proceedings.mlr.press/v80/balle18a.html}
}

@inproceedings{ho2020denoising,
  title     = {Denoising Diffusion Probabilistic Models},
  author    = {Ho, J. and Jain, A. and Abbeel, P.},
  booktitle = {Advances in Neural Information Processing Systems},
  volume    = {33},
  pages     = {6840-6851},
  year      = {2020}
}

@inproceedings{rombach2022high,
  title     = {High-Resolution Image Synthesis with Latent Diffusion Models},
  author    = {Rombach, R. and Blattmann, A. and Lorenz, D.
               and Esser, P. and Ommer, B.},
  booktitle = {Proceedings of the IEEE/CVF Conference on Computer Vision
               and Pattern Recognition},
  pages     = {10684-10695},
  year      = {2022},
  doi       = {10.1109/CVPR52688.2022.01042}
}

@article{fedmekt2025,
  title   = {FedMEKT: Distillation-Based Embedding Knowledge Transfer
             for Multimodal Federated Learning},
  author  = {Le, H.Q. and Nguyen, M.N.H. and Thwal, C.M.
             and Qiao, Y. and Zhang, C. and Hong, C.S.},
  journal = {Neural Networks},
  volume  = {183},
  pages   = {107017},
  year    = {2025},
  doi     = {10.1016/j.neunet.2024.107017}
}

@article{fedmepd2025,
  title   = {Federated Modality-Specific Encoders and Partially Personalized
             Fusion Decoder for Multimodal Brain Tumor Segmentation},
  author  = {Liu, H. and Wei, D. and Dai, Q. and Wu, X.
             and Zheng, Y. and Wang, L.},
  journal = {Medical Image Analysis},
  volume  = {106},
  pages   = {103759},
  year    = {2025},
  doi     = {10.1016/j.media.2025.103759}
}

@article{torfi2022dpgan,
  title   = {Differentially Private Synthetic Medical Data Generation
             Using Convolutional {GANs}},
  author  = {Torfi, A. and Fox, E.A. and Reddy, C.K.},
  journal = {Information Sciences},
  volume  = {586},
  pages   = {485-500},
  year    = {2022},
  doi     = {10.1016/j.ins.2021.12.018}
}

@article{niehues2024histology,
  title   = {Using Histopathology Latent Diffusion Models as
             Privacy-Preserving Dataset Augmenters Improves Downstream
             Classification Performance},
  author  = {Niehues, J.M. and others},
  journal = {Computers in Biology and Medicine},
  volume  = {175},
  pages   = {108410},
  year    = {2024},
  doi     = {10.1016/j.compbiomed.2024.108410}
}

@article{ISLAM2026129418,
title = {Study of thermo-fluid dynamics of magnetised hybrid nanofluids over a Bi-directional surface using ANN-optimization},
author = {Islam, T. and Rana, B.J. and Ali, M.Y. and Hossain, K.E. and Mukherjee, A. and Hossain, S.N. and Afikuzzaman, M.},
journal = {Journal of Molecular Liquids},
volume = {449},
pages = {129418},
year = {2026},
issn = {0167-7322},
url= {https://www.sciencedirect.com/science/article/pii/S0167732226001881}}

@article{islam2025artificial,
  title={Artificial Neural Network Modeling of Magnetic Nanoparticle-Enhanced Sisko Blood Nanofluid Flow over an Inclined Stretching Surface with Non-Uniform Heating and Thermophoretic Effects},
  author={Islam, T. and Rana, B.J. and Ali, M.Y. and Hossain, K.E. and Mukherjee, A. and Islam, S. and Afikuzzaman, M.},
  journal={International Journal of Thermofluids},
  pages={101542},
  year={2025},
  publisher={Elsevier},
  doi = {https://doi.org/10.1016/j.ijft.2025.101542}
}

@article{rana2025neural,
  title={Neural network-based computational evaluation of periodic electroosmotic flow in propylene glycol-water ternary nanofluids with oxytactic microbes},
  author={Rana, B.J. and Islam, T. and Ali, M.Y. and Islam, S. and Hossain, K.E. and Mukherjee, A. and Islam, M.R. and Afikuzzaman, M.},
  journal={Journal of Molecular Liquids},
  pages={128593},
  year={2025},
  publisher={Elsevier},
  doi = {10.1016/j.molliq.2025.128593}
}

@article{vaswani2017attention,
  title     = {Attention Is All You Need},
  author    = {Vaswani, A. and Shazeer, N. and Parmar, N.
               and Uszkoreit, J. and Jones, L. and Gomez, A.N.
               and Kaiser, L. and Polosukhin, I.},
  journal = {Advances in Neural Information Processing Systems},
  volume    = {30},
  year      = {2017}
}

@article{kuhn1955hungarian,
  title   = {The Hungarian Method for the Assignment Problem},
  author  = {Kuhn, H.W.},
  journal = {Naval Research Logistics Quarterly},
  volume  = {2},
  number  = {1-2},
  pages   = {83-97},
  year    = {1955},
  doi     = {10.1002/nav.3800020109}
}

@misc{ho2022classifierfree,
  title         = {Classifier-Free Diffusion Guidance},
  author        = {Ho, J. and Salimans, T.},
  year          = {2022},
  eprint        = {2207.12598},
  archivePrefix = {arXiv},
  primaryClass  = {cs.LG},
  doi           = {10.48550/arXiv.2207.12598}
}

@article{yang2023medmnist,
  title   = {{MedMNIST v2}: A Large-Scale Lightweight Benchmark for 2D and 3D Biomedical Image Classification},
  author  = {Yang, J. and Shi, R. and Wei, D. and Liu, Z. and Zhao, L. and Ke, B. and Pfister, H. and Ni, B.},
  journal = {Scientific Data},
  volume  = {10},
  number  = {1},
  pages   = {41},
  year    = {2023},
  doi     = {10.1038/s41597-022-01721-8}
}

@article{goldberger2000physionet,
  title   = {{PhysioBank, PhysioToolkit, and PhysioNet}: Components of a New Research Resource for Complex Physiologic Signals},
  author  = {Goldberger, A.L. and Amaral, L.A.N. and Glass, L. and Hausdorff, J.M. and Ivanov, P.C. and Mark, R.G. and Mietus, J.E. and Moody, G.B. and Peng, C. and Stanley, H.E.},
  journal = {Circulation},
  volume  = {101},
  number  = {23},
  pages   = {e215-e220},
  year    = {2000},
  doi     = {10.1161/01.CIR.101.23.e215}
}

@article{Tomarchio2025,
  author  = {Tomarchio, S.D. and Punzo, A. and Ferreira, J.T.},
  title   = {A New Look at the Dirichlet Distribution: Robustness, Clustering, and Both Together},
  journal = {Journal of Classification},
  year    = {2025},
  volume  = {42},
  pages   = {31-53},
  doi     = {10.1007/s00357-024-09480-4}
}

@article{ioffe2015batch,
  title     = {Batch Normalization: Accelerating Deep Network Training
               by Reducing Internal Covariate Shift},
  author    = {Ioffe, S. and Szegedy, C.},
  journal = {Proceedings of the 32nd International Conference on
               Machine Learning},
  series    = {Proceedings of Machine Learning Research},
  volume    = {37},
  pages     = {448-456},
  year      = {2015},
  publisher = {PMLR}
}

@article{hendrycks2016gelu,
  title         = {Gaussian Error Linear Units ({GELU}s)},
  author        = {Hendrycks, D. and Gimpel, K.},
  journal       = {arXiv preprint arXiv:1606.08415},
  year          = {2016},
  eprint        = {1606.08415},
  archivePrefix = {arXiv},
  primaryClass  = {cs.LG}
}

@article{lin2014network,
  title     = {Network in Network},
  author    = {Lin, M. and Chen, Q. and Yan, S.},
  journal = {International Conference on Learning Representations},
  year      = {2014}
}

@article{ba2016layer,
  title         = {Layer Normalization},
  author        = {Ba, J.L. and Kiros, J.R.
                   and Hinton, G.E.},
  journal       = {arXiv preprint arXiv:1607.06450},
  year          = {2016},
  eprint        = {1607.06450},
  archivePrefix = {arXiv},
  primaryClass  = {stat.ML}
}

@article{song2021ddim,
  title     = {Denoising Diffusion Implicit Models},
  author    = {Song, J. and Meng, C. and Ermon, S.},
  journal = {International Conference on Learning Representations},
  year      = {2021}
}

@article{tharwat2021classification,
  title   = {Classification Assessment Methods},
  author  = {Tharwat, A.},
  journal = {Applied Computing and Informatics},
  volume  = {17},
  number  = {1},
  pages   = {168-192},
  year    = {2021},
  doi     = {10.1016/j.aci.2018.08.003}
}

@article{fawcett2006roc,
  title   = {An Introduction to {ROC} Analysis},
  author  = {Fawcett, T.},
  journal = {Pattern Recognition Letters},
  volume  = {27},
  number  = {8},
  pages   = {861-874},
  year    = {2006},
  doi     = {10.1016/j.patrec.2005.10.010}
}

@article{saito2015precision,
  title   = {The Precision--Recall Plot Is More Informative than the
             {ROC} Plot When Evaluating Binary Classifiers on
             Imbalanced Datasets},
  author  = {Saito, Takaya and Rehmsmeier, Marc},
  journal = {PLOS ONE},
  volume  = {10},
  number  = {3},
  pages   = {e0118432},
  year    = {2015},
  doi     = {10.1371/journal.pone.0118432}
}

@article{brodersen2010balanced,
  title     = {The Balanced Accuracy and Its Posterior Distribution},
  author    = {Brodersen, Kay H. and Ong, Cheng Soon and Stephan,
               Klaas E. and Buhmann, Joachim M.},
  journal = {Proceedings of the 20th International Conference on
               Pattern Recognition},
  pages     = {3121-3124},
  year      = {2010},
  publisher = {IEEE},
  doi       = {10.1109/ICPR.2010.764}
}

@article{matthews1975comparison,
  title   = {Comparison of the Predicted and Observed Secondary
             Structure of {T4} Phage Lysozyme},
  author  = {Matthews, B.W.},
  journal = {Biochimica et Biophysica Acta (BBA) - Protein Structure},
  volume  = {405},
  number  = {2},
  pages   = {442-451},
  year    = {1975},
  doi     = {10.1016/0005-2795(75)90109-9}
}

@article{brier1950verification,
  title   = {Verification of Forecasts Expressed in Terms of Probability},
  author  = {Brier, G.W.},
  journal = {Monthly Weather Review},
  volume  = {78},
  number  = {1},
  pages   = {1--3},
  year    = {1950},
  doi     = {10.1175/1520-0493(1950)078<0001:VOFEIT>2.0.CO;2}
}

@article{guo2017calibration,
  title     = {On Calibration of Modern Neural Networks},
  author    = {Guo, C. and Pleiss, G. and Sun, Y. and Weinberger,
               K.Q.},
  journal = {Proceedings of the 34th International Conference on
               Machine Learning},
  series    = {Proceedings of Machine Learning Research},
  volume    = {70},
  pages     = {1321-1330},
  year      = {2017},
  publisher = {PMLR}
}

@inproceedings{heusel2017gans,
  title     = {{GANs} Trained by a Two Time-Scale Update Rule Converge
               to a Local {Nash} Equilibrium},
  author    = {Heusel, Martin and Ramsauer, Hubert and Unterthiner,
               Thomas and Nessler, Bernhard and Hochreiter, Sepp},
  booktitle = {Advances in Neural Information Processing Systems},
  volume    = {30},
  pages     = {6626-6637},
  year      = {2017}
}

@article{gretton2012kernel,
  title   = {A Kernel Two-Sample Test},
  author  = {Gretton, A. and Borgwardt, K.M. and Rasch,
             M.J. and Sch{\"o}lkopf, B. and Smola,
             A.J.},
  journal = {Journal of Machine Learning Research},
  volume  = {13},
  number  = {25},
  pages   = {723-773},
  year    = {2012}
}

@article{naeem2020reliable,
  title     = {Reliable Fidelity and Diversity Metrics for Generative
               Models},
  author    = {Naeem, M.F. and others},
  journal = {Proceedings of the 37th International Conference on
               Machine Learning},
  series    = {Proceedings of Machine Learning Research},
  volume    = {119},
  pages     = {7176-7185},
  year      = {2020},
  publisher = {PMLR}
}

@article{lecun1998gradient,
  title   = {Gradient-Based Learning Applied to Document Recognition},
  author  = {LeCun, Y. and Bottou, L. and Bengio, Y. and Haffner, P.},
  journal = {Proceedings of the IEEE},
  volume  = {86},
  number  = {11},
  pages   = {2278-2324},
  year    = {1998},
  doi     = {10.1109/5.726791}
}

\end{document}